\pdfoutput=1
\documentclass[preprints,article,accept,pdftex,moreauthors]{Definitions/mdpi}
\usepackage{makecell}
\usepackage{graphicx}
\usepackage{booktabs}
\usepackage{tabularx}
\firstpage{1} 
\pubvolume{1}
\issuenum{1}
\articlenumber{0}
\pubyear{2026}
\copyrightyear{2026}
\externaleditor{Firstname Lastname} 
\datereceived{2 August 2026} 
\daterevised{7 September 2026} 
\dateaccepted{11 September 2026} 
\datepublished{ } 

\Title{DTKDP: A Dual Teacher Knowledge Distillation and Pruning Framework for Lightweight Oriented SAR Ship Detection}

\Author{Yuming 
 Li *, Fan Zhang and Alin M. Achim}

\AuthorNames{Yuming Li, Fan Zhang and Alin M. Achim}

\address[1]{%
Visual Information Labs, University of Bristol,
Bristol BS1 6AZ, UK;
fan.zhang@bristol.ac.uk (F.Z.);
alin.achim@bristol.ac.uk (A.M.A.)}

\corres{\hangafter=1 \hangindent=1.05em \hspace{-0.82em}Correspondence: yuming.li@bristol.ac.uk}

\abstract{
Two-stage oriented detectors achieve high localization accuracy in {synthetic aperture radar (SAR)} ship
detection, but their large backbones, feature pyramids, proposal modules, and
heavy region of interest (RoI) heads hinder deployment. Existing lightweight
SAR ship detectors typically use one-stage frameworks that lack proposal-level
refinement for precise rotated localization. This paper presents a {dual-teacher knowledge distillation and pruning} (DTKDP) framework for lightweight oriented SAR ship detection. DTKDP introduces learnable gates into convolutional, normalization, and linear layers to prune convolutional channels and RoI-head neurons. Rotated proposal alignment (RPA) distills teacher and student predictions in a shared teacher-generated rotated proposal space, while a dual-teacher scheme combines classification and regression guidance
from a homogeneous main teacher with complementary classification cues from a
heterogeneous auxiliary teacher. Experiments on the {SAR Ship Detection Dataset (SSDD)} and {Rotated Ship Detection Dataset in SAR Images (RSDD-SAR)} show that DTKDP reduces the parameters of {Oriented Region-based Convolutional Neural Network (Oriented R-CNN)} and RoI Transformer equipped with ResNet-50 backbones by 87.5--91.8\% and their floating-point operations (FLOPs) by 75.6--79.9\%. In terms of {average precision (AP)} and {mean average precision (mAP)}, the resulting Oriented R-CNN-slim and RoI Transformer-slim retain accuracy close to their full-scale counterparts. Relative changes across AP$_{50}$, AP$_{75}$, mAP$_{50:75}$, and mAP$_{50:95}$ range from a 2.38\% decrease to a 0.65\% improvement. Compared with RTMDet-tiny, they improve all four metrics on both datasets by 0.52--27.55\% and consistently surpass representative distillation methods, demonstrating a favorable accuracy--efficiency trade-off.
}


\addhighlights{yes}
\renewcommand{\addhighlights}{%

\noindent\textbf{What are the main findings?}
\begin{itemize}[labelsep=2.5mm,topsep=-3pt]
\item We propose a component-wise pruning strategy that compresses the entire
two-stage oriented detection pipeline, including the parameter-intensive
detection head, reducing the number of model parameters and floating-point
operations by up to 91.8\% and 79.9\%, respectively.

\item We develop Dual Teacher Knowledge Distillation (DTKD), which uses
Rotated Proposal Alignment (RPA) to align teacher and student predictions on
teacher-generated rotated proposals and combines supervision from the main
and auxiliary teachers to improve knowledge transfer after pruning.
\end{itemize}

\vspace{3pt}

\noindent\textbf{What are the implications of the main findings?}
\begin{itemize}[labelsep=2.5mm,topsep=-3pt]
\item The substantial reductions in model size and computation make two-stage SAR ship detectors more practical for resource-constrained onboard and edge platforms.

\item The results demonstrate that two-stage architectures can be substantially compressed for SAR ship detection while maintaining strong localization accuracy.

\end{itemize}
}
\begin{document}


\section{Introduction}
\label{sec:intro}

{Synthetic aperture radar (SAR) is well suited to large-scale maritime monitoring, with~spaceborne systems capable of imaging swaths spanning tens to hundreds of kilometers while providing day-and-night and all-weather observation capabilities~\cite{asiyabi2023sar}.} Ship detection is one of its most
prominent applications, with~important roles in maritime surveillance,
traffic monitoring, and~security~\cite{li2022deep}. Conventional
approaches have largely relied on handcrafted features and statistical
decision rules, including constant false alarm rate (CFAR) detectors
based on local clutter estimation and adaptive thresholding. A~range of
CFAR variants has been developed to preserve target responses, suppress
clutter interference, reduce contamination from neighboring targets,
and adapt reference regions to local scene structures~\cite{hou2015multilayer,leng2015bilateral,tao2016robust,
ai2021robust,pappas2018superpixel}. {Advanced target-detection methods for high-frequency surface wave radar (HFSWR), including time--frequency and high-resolution processing approaches, have also been investigated to improve maritime target detection and
localization~\cite{yang2021joint,cai2021ship,golubovic2024high}.} Despite these advances, the~performance of CFAR methods may remain sensitive to clutter-model
assumptions, reference-region construction, and~threshold selection,
particularly in heterogeneous inshore and dense multi-target~scenes.

More recently, {convolutional neural network (CNN)-based} detectors have been developed to learn
discriminative and multiscale representations from SAR imagery.
Representative early CNN-based methods for SAR maritime target and
ship detection employed horizontal bounding boxes for target
localization~\cite{ma2018ship,zhao2020attention}. For~slender, densely distributed, and~arbitrarily oriented ships, horizontal bounding boxes may include substantial background regions and overlap considerably in crowded scenes. In~contrast, oriented bounding boxes better capture the geometry
and orientation of slender vessels and alleviate box overlap in densely
distributed scenes, making them more suitable for accurate SAR ship
localization~\cite{he2021polar,fpddet2023}.

Among existing oriented object detection architectures, two-stage detectors, such as the region-of-interest (RoI) Transformer~\cite{Ding_2019_CVPR} and {Oriented Region-based Convolutional Neural Network (Oriented R-CNN)}~\cite{Xie_2021_ICCV}, have demonstrated strong localization performance. These methods first generate candidate regions and
then refine them at the RoI level for classification and rotated box regression. The~RoI-level alignment and refinement steps help reduce the mismatch between coarse proposals and oriented objects, which is especially useful for dense or arbitrarily oriented targets. However, this two-stage RoI-based pipeline incurs additional computational and memory costs. RoI feature extraction and alignment, per-proposal processing, and~fully connected detection heads introduce overhead that increases with the number of candidate regions. In~particular, the~heavy detection head has been recognized as an efficiency bottleneck in two-stage detectors~\cite{li2017lighthead}. As~a result, directly deploying such detectors on resource-constrained platforms, such as onboard satellite systems or edge-based maritime monitoring devices, remains challenging~\cite{garcia2024onboard,wiehle2021onboard}.

A common strategy for efficient SAR ship detection is to adopt compact one-stage pipelines derived from real-time detectors such as {You Only Look Once (YOLO)}~\cite{yolo} and RTMDet~\cite{lyu2022rtmdet}. For~oriented SAR ship detection, some recent lightweight models further adapt this paradigm through compact feature aggregation and rotation-aware prediction heads~\cite{lsrdet,rsabmnet,zhang2025rtmdet_sar}. Although~efficient, these dense predictors discard proposal-level refinement. In~cluttered SAR scenes with dense or highly elongated ships, small angular and boundary errors can cause large drops in rotated intersection over union (IoU), especially under strict localization thresholds. This motivates a complementary direction that compresses accurate two-stage oriented detectors to a deployable scale while preserving their proposal-based localization advantage. Structured pruning provides a principled way to compress overparameterized networks by removing redundant channels or filters~\cite{wen2016learning,luo2017thinet,he2017channel,liu2017network}. However, pruning a two-stage oriented detector involves more than compressing its backbone. Conventional channel and filter pruning primarily focuses on convolutional structures, whereas redundancy in a two-stage detector is distributed across multiple coupled components, including the backbone, feature pyramid network (FPN), region proposal network (RPN), and~RoI head. Consequently, pruning only convolutional structures does not directly reduce the fully connected neurons in the RoI head, even though this component can account for a substantial share of the model parameters. Effective compression of a two-stage oriented detector therefore requires structural pruning to cover the complete pipeline, from~feature extraction and proposal generation to RoI~refinement.

To improve the performance of lightweight models, knowledge distillation~\cite{hinton2015distilling,li2017mimicking,chen2017learning} has emerged as an effective way to train compact object detectors. Existing detection distillation methods mainly rely
on feature imitation or prediction mimicking~\cite{shu2021cwd,cao2022pkd,yang2022mgd,zheng2022localization,wang2024crosskd}.
However, both paradigms require suitable teacher--student correspondence.
Feature imitation depends on aligned intermediate representations, whereas prediction mimicking requires a well-defined correspondence between the outputs being compared. This requirement becomes restrictive for heavily pruned two-stage oriented detectors, where the student may differ from the teacher in backbone channels, feature-pyramid channels, proposal modules, and~RoI-head neuron widths. Moreover, a~single teacher may not provide uniformly strong supervision for ships with diverse scales and aspect ratios. Multi-teacher distillation has been explored in image super-resolution to aggregate knowledge from multiple teacher models~\cite{yao2022mtkdsr,jiang2024mtkd}, but~its application to
proposal-based oriented SAR ship detection remains~underexplored.

In this context, this paper proposes a {dual-teacher knowledge distillation and pruning (DTKDP)} framework for lightweight oriented SAR ship detection. {Unlike existing lightweight SAR ship detectors that mainly achieve efficiency by redesigning one-stage dense architectures, DTKDP takes established high-accuracy two-stage oriented detectors as compression targets while retaining their proposal-based refinement paradigm. Previous SAR ship detection studies have also combined pruning with knowledge distillation to construct compact detectors~\cite{chen2021learning_slimming_sar,hu2025lightweight_sar,xu2026lightweight_framework}. These methods are mainly developed for one-stage dense detectors, where pruning primarily targets convolutional structures and knowledge is typically transferred from a single teacher through feature or prediction distillation. In~contrast, DTKDP extends structural pruning to the complete two-stage oriented detection pipeline, including the fully connected RoI head, introduces Rotated Proposal Alignment (RPA) to establish reliable proposal-level teacher--student correspondence, and~employs a dual-teacher strategy that combines a homogeneous main teacher with a heterogeneous auxiliary teacher. The~main teacher provides classification and rotated box regression supervision, while the auxiliary teacher contributes complementary classification knowledge.} The main contributions are summarized as follows:

\begin{itemize}
    \item We address the underexplored problem of compressing full two-stage
    oriented SAR ship detectors, rather than redesigning lightweight one-stage
    dense predictors~\cite{lsrdet, rsabmnet, zhang2025rtmdet_sar}. The~proposed {component-wise pruning strategy 
} jointly
    compresses convolutional channels and RoI-head neurons, thereby covering
    the backbone, feature pyramid, proposal module, and~detection head within a
    unified pruning~framework.

    \item We propose {Rotated Proposal Alignment} to address the proposal mismatch
    problem in distilling pruned two-stage oriented detectors. Different from
    conventional feature imitation~\cite{li2017mimicking,shu2021cwd,cao2022pkd,yang2022mgd,yang2022fgd} or existing prediction-matching approaches~\cite{chen2017learning, zheng2022localization,wang2024crosskd}, RPA compares teacher and student predictions in a shared teacher-generated
    rotated proposal space, providing reliable prediction-level supervision after
    pruning.

    \item We develop a {dual-teacher distillation strategy} to leverage complementary teacher knowledge for SAR ships with diverse scales and aspect ratios. The~homogeneous teacher is used for stable classification and localization transfer, while the heterogeneous teacher provides complementary
    classification knowledge, enabling accuracy recovery without directly mixing incompatible regression outputs.
\end{itemize}

\begin{figure}[H]
\centering

\includegraphics[
    width=1.0\textwidth
]{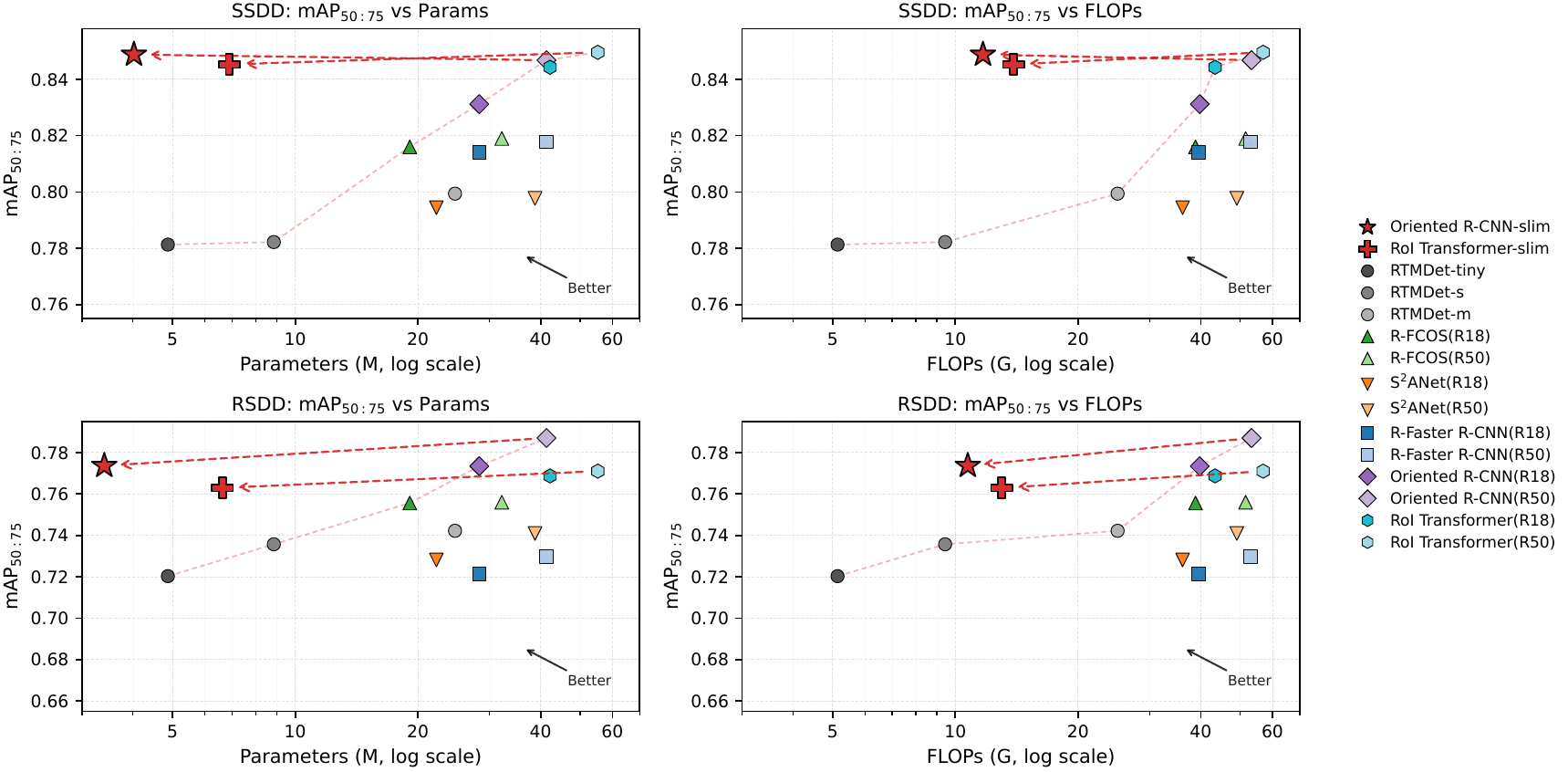}

\caption{
Accuracy--efficiency trade-off on SSDD and RSDD-SAR datasets.
Each point denotes an oriented detector benchmarked in our experiment,
with the shared legend shown on the right.
The x-axis is shown in log scale for readability.
The proposed slim two-stage detectors achieve competitive
mAP$_{50:75}$ with substantially fewer parameters and FLOPs.
}

\label{fig:teaser_tradeoff}
\end{figure}

Experiments on the {SAR Ship Detection Dataset (SSDD)}~\cite{zhang2021ssdd} and {Rotated Ship Detection Dataset in SAR Images (RSDD-SAR)}~\cite{xu2022rsddsar}
validate DTKDP when it is applied to Oriented R-CNN and RoI Transformer
equipped with ResNet-50 (R50)~\cite{he2016deep} backbones. It reduces the
parameters of Oriented R-CNN (R50) by 90.3\% and 91.8\% on SSDD and RSDD-SAR,
respectively, and~those of RoI Transformer (R50) by 87.5\% and 88.1\%. It also
reduces floating-point operations (FLOPs) by 78.1\% and 79.9\% for Oriented R-CNN (R50), and~by 75.6\% and
77.2\% for RoI Transformer (R50). Despite the substantial complexity reduction, the~resulting slim models, denoted Oriented R-CNN-slim and RoI Transformer-slim, maintain strong rotated localization performance. In~terms of {average precision (AP)} and {mean average precision (mAP)}, the~relative changes across AP$_{50}$, AP$_{75}$, mAP$_{50:75}$, and~mAP$_{50:95}$ range from a 2.38\% decrease to a 0.65\% improvement compared with their full-scale counterparts. Compared with RTMDet-tiny, the~compressed detectors improve all four accuracy metrics on both datasets, with~relative gains ranging from 0.52\% to 27.55\%, and~consistently outperform representative distillation baselines~\cite{shu2021cwd,cao2022pkd,yang2022mgd}. The~excellent complexity--performance trade-off of the proposed method is illustrated by Figure \ref{fig:teaser_tradeoff}.

The remainder of this paper is organized as follows.
\Cref{sec:related_work} reviews related work on oriented object
detection, lightweight SAR ship detection, and~knowledge distillation.
\Cref{sec:methods} presents the proposed DTKDP framework, including
component-wise pruning, Rotated Proposal Alignment, and~dual-teacher knowledge
distillation. \Cref{sec:results} describes the experimental setup and presents
analyses of component-wise pruning, teacher complementarity, and distillation
loss-weight sensitivity, followed by comparisons with representative methods,
distillation ablation studies, auxiliary-teacher comparison,
and~qualitative results. \Cref{sec:discussion} discusses the findings and limitations. Finally, \Cref{sec:conclusions} concludes the~paper.

\vspace{-3pt}
\unskip

\section{Related~Work}
\label{sec:related_work}
\unskip
\subsection{Oriented Object~Detection}
Oriented object detection has been widely studied in remote sensing imagery,
where objects often appear with arbitrary orientations, large aspect-ratio
variations, and~dense layouts. Representative methods can be organized along
several architectural lines, including two-stage, one-stage, transformer-based,
and sparse-proposal approaches. Two-stage detectors, including RoI
Transformer~\cite{Ding_2019_CVPR}, Oriented R-CNN~\cite{Xie_2021_ICCV}, and~ReDet~\cite{han2021redet}, emphasize proposal quality and RoI-level refinement.
One-stage detectors, such as S$^2$A-Net~\cite{han2022s2anet},
R3Det~\cite{yang2021r3det}, and~Rotated FCOS~\cite{fcosr}, perform oriented
detection through dense prediction without RoI-level refinement.
Transformer-based methods, including Rotated-DETR~\cite{kim2023rotateddetr}
and ARS-DETR~\cite{zeng2024arsdetr}, further explore end-to-end set prediction
for oriented object detection. In~SAR ship detection, sparse-proposal methods,
such as Sparse R-CNN OBB~\cite{kamirul2025sparsercnobb} and R-Sparse
R-CNN~\cite{kamirul2025rsparsercnobb}, introduce learnable oriented proposals to
reduce the dependence on dense candidates. Although~these methods have advanced
oriented object detection, high-accuracy detection pipelines often remain
computationally~demanding.

\subsection{Lightweight SAR Ship~Detection}
Efficient SAR ship detection is important for onboard and edge deployment,
where computation, memory, and~power budgets are limited. Existing efficient
detectors are largely based on one-stage real-time frameworks, such as the
YOLO family and RTMDet~\cite{yolo,lyu2022rtmdet}. Building on these
frameworks, lightweight SAR ship detectors typically improve efficiency
through compact backbones, lightweight feature aggregation, or~simplified
prediction heads, including LMSD-YOLO~\cite{lmsdyolo},
LH-YOLO~\cite{lhyolo}, LD-YOLO~\cite{ldyolo}, and~YOLOv7oSAR~\cite{yolov7osar}. Recent oriented variants further incorporate
rotation-aware prediction and lightweight feature enhancement, including
LSR-Det~\cite{lsrdet}, R-SABMNet~\cite{rsabmnet}, and~an RTMDet-based
arbitrary-direction ship detector~\cite{zhang2025rtmdet_sar}. At~the system
level, CFAR-based candidate screening has also been combined with YOLOv4-tiny
to enable near-real-time onboard SAR ship detection on an embedded
{graphics processing unit (GPU)}~\cite{xu2021onboard}.
{Beyond architectural redesign, pruning and knowledge distillation have
also been combined to further reduce detector complexity. Tiny
YOLO-Lite~\cite{chen2021learning_slimming_sar} combines channel-level pruning
with knowledge distillation, while Hu and
Miao~\cite{hu2025lightweight_sar} integrate lightweight architectural design,
network pruning, and~knowledge distillation. LMFAN~\cite{shi2025lmfan}
combines channel pruning with channel-wise feature distillation, whereas
LGNet~\cite{chen2025lgnet} incorporates lightweight architectural design,
structured pruning, and~prediction-level distillation. More recently, Xu~et~al.~\cite{xu2026lightweight_framework} combined layerwise pruning with
dual-stream distillation. These representative pruning--distillation
approaches are predominantly developed within one-stage pipelines and mainly
reduce convolutional complexity. In~contrast, two-stage oriented detectors
additionally involve proposal generation and parameter-intensive fully
connected RoI heads, motivating structural pruning beyond conventional
convolutional components.}

\subsection{Knowledge Distillation for Compact~Detectors}
Knowledge distillation (KD) transfers knowledge from a high-capacity teacher
to a compact student through softened output
distributions~\cite{hinton2015distilling}. Extending KD to object detection is
challenging because classification and localization must be learned jointly.
Early work combined classification distillation, box regression, and~intermediate feature supervision~\cite{chen2017learning}. Feature-based
methods subsequently transferred informative intermediate representations
through region-aware, channel-wise, correlation-based, or~reconstruction-based
objectives~\cite{li2017mimicking,wang2019fine-grained,
du2021feature_richness,shu2021cwd,cao2022pkd,yang2022fgd,yang2022mgd}.
Prediction-oriented methods instead transfer knowledge closer to detector
outputs. Localization Distillation~\cite{zheng2022localization} transfers
probabilistic localization knowledge, while CrossKD~\cite{wang2024crosskd}
matches cross-head predictions generated using the teacher head. However,
many methods assume compatible teacher--student representations or directly
comparable detector outputs. After~full-component pruning, structural
differences between the teacher and student complicate feature- and head-level
knowledge transfer. {Table~\ref{tab:related_comparison} provides a concise comparison of representative pruning--distillation approaches and highlights the methodological gap addressed by DTKDP.}

\begin{table}[H]
\tiny
\centering
\setlength{\tabcolsep}{7pt}
\renewcommand{\arraystretch}{1.08}

\caption{
Comparison of representative pruning--distillation approaches for lightweight
SAR ship detection. HBB denotes horizontal bounding boxes, and OBB denotes
oriented bounding boxes. Conv. denotes convolutional layers, and FC denotes
fully connected layers. ``--'' denotes not applicable, ``$\checkmark$''
indicates that the corresponding mechanism is explicitly included, and
``NR'' denotes that the hardware platform was not reported.
}
\label{tab:related_comparison}

\makebox[\textwidth][c]{%
\begin{tabularx}{1.10\textwidth}{
@{}
>{\raggedright\arraybackslash}p{2.25cm}
*{6}{>{\centering\arraybackslash}X}
>{\centering\arraybackslash}p{1.55cm}
>{\centering\arraybackslash}p{2.00cm}
>{\centering\arraybackslash}p{1.85cm}
@{}
}

\toprule

\textbf{Approach} &
\makecell{\textbf{Detector}\\\textbf{Paradigm}} &
\makecell{\textbf{Box}\\\textbf{Type}} &
\makecell{\textbf{Pruning}\\\textbf{Scope}} &
\makecell{\textbf{KD}\\\textbf{Target}} &
\makecell{\textbf{Teacher}\\\textbf{Setting}} &
\makecell{\textbf{Proposal}\\\textbf{Alignment}} &
\makecell{\textbf{RoI-Head}\\\textbf{Compression}} &
\makecell{\textbf{Deployment}\\\textbf{Metrics}} &
\textbf{Hardware} \\

\midrule

Tiny YOLO-Lite~\cite{chen2021learning_slimming_sar}
& One-stage
& HBB
& Conv.
& \makecell{Feature +\\prediction}
& Single
& --
& --
& \makecell{Params/FLOPs/\\size/FPS}
& GTX 1080 Ti \\

Hu and Miao~\cite{hu2025lightweight_sar}
& One-stage
& HBB
& Conv.
& \makecell{Feature +\\prediction}
& Single
& --
& --
& \makecell{Params/FLOPs/\\size/time}
& RTX 3090 \\

LMFAN~\cite{shi2025lmfan}
& One-stage
& HBB
& Conv.
& Feature
& Single
& --
& --
& \makecell{Params/\\complexity}
& \makecell{RTX 4070 Ti\\Super} \\

LGNet~\cite{chen2025lgnet}
& One-stage
& HBB/OBB
& Conv.
& Prediction
& Single
& --
& --
& \makecell{Params/FLOPs/\\FPS/latency}
& \makecell{RTX 2080 Ti /\\Atlas AIpro-20T} \\

Xu et al.~\cite{xu2026lightweight_framework}
& One-stage
& HBB/OBB
& Conv.
& \makecell{Feature +\\prediction}
& Single
& --
& --
& \makecell{Params/FLOPs/\\FPS}
& NR \\

\midrule

\multicolumn{10}{@{}l@{}}{Ours} \\
\addlinespace[2pt]

DTKDP
& Two-stage
& OBB
& Conv. + FC
& Prediction
& Dual
& $\checkmark$ (RPA)
& $\checkmark$ (FC)
& \makecell{Params/FLOPs/\\size/memory/FPS}
& RTX 3090 \\

\bottomrule

\end{tabularx}%
}

\end{table}

\section{Dual-Teacher Knowledge Distillation and Pruning~Framework}
\label{sec:methods}

The proposed framework consists of two phases. First, learnable gates are
introduced into convolutional, normalization, and~linear layers to enable
component-wise structured pruning of the full two-stage oriented detector.
Second, the~pruned detector is trained using dual-teacher knowledge distillation.
The following subsections describe the model compression strategy and the
dual-teacher distillation framework, including the teacher configuration,
RPA mechanism, and~distillation~objective.

\subsection{Model~Compression}
\begin{figure}[H]
\centering

\makebox[\textwidth][c]{%
    \includegraphics[
        width=1.25\textwidth
    ]{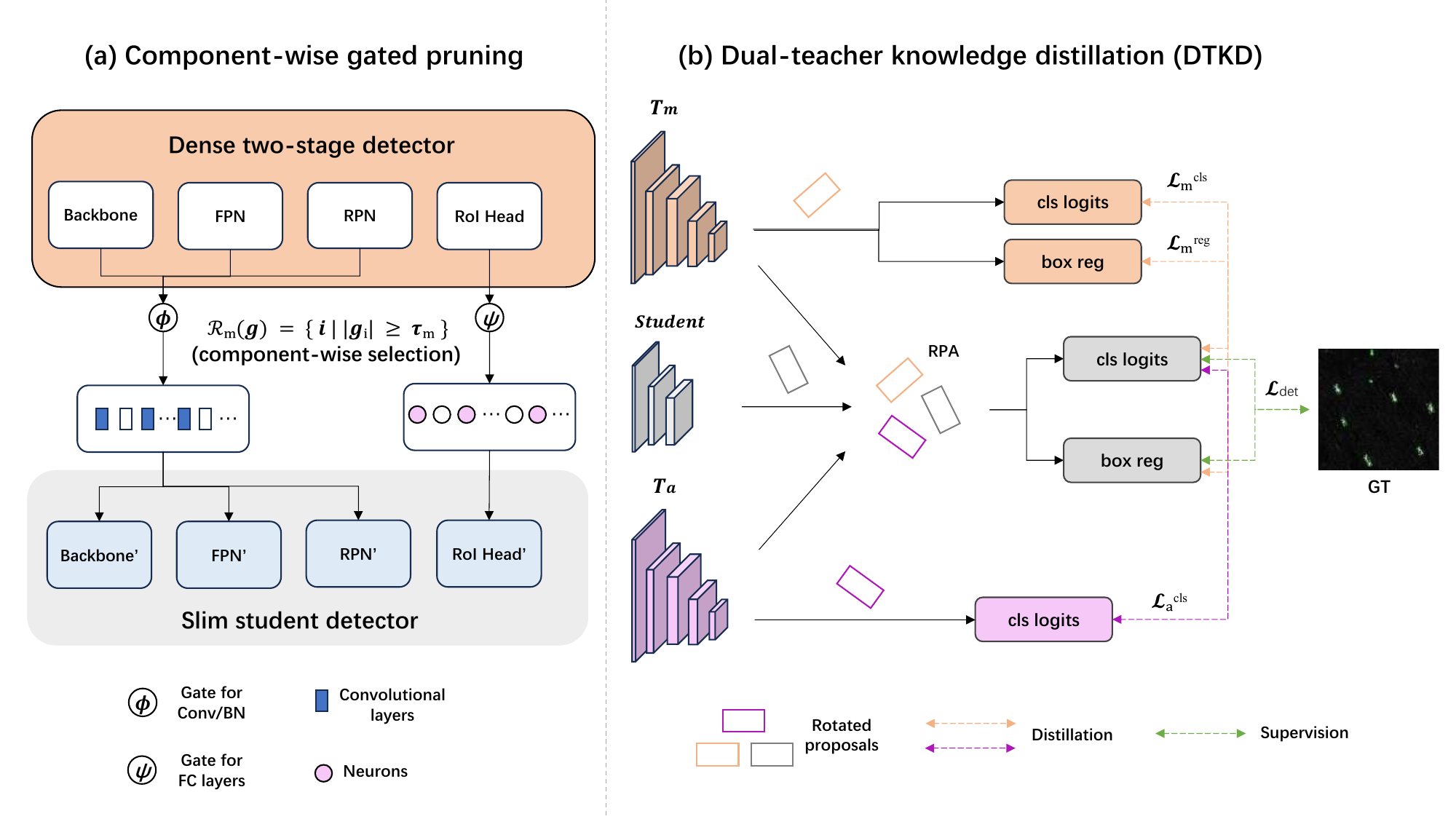}
}

\caption{
Overview of the proposed pruning-distillation framework.
(\textbf{a}) Component-wise gated pruning compresses the dense two-stage detector
by pruning convolutional channels and RoI-head neurons.
(\textbf{b}) Dual-teacher knowledge distillation trains the student with
ground-truth supervision and prediction-level guidance in aligned rotated
proposal spaces, where the main teacher provides classification and regression
supervision and the auxiliary teacher provides complementary classification cues.
}

\label{fig:dtkd_framework}
\end{figure}

As illustrated in Figure \ref{fig:dtkd_framework}a, two-stage oriented detectors typically contain heterogeneous components, including the backbone, FPN, RPN, and~RoI head. These components play different roles and exhibit different pruning sensitivities, motivating a component-wise pruning strategy
tailored to these modules. We therefore introduce learnable gates to estimate structural importance and perform component-wise structured pruning of both convolutional channels and linear-layer neurons. Specifically, we first reparameterize convolutional, normalization, and~linear layers with removable gates, then perform sparsity training on the gated model initialized from a pretrained detector, and~finally derive a compact detector by pruning low-importance structures within each~component.

\subsubsection{Gate-Based Layer~Reparameterization}

\noindent\textbf{Gated Convolution and Batch Normalization. 
}
Following the gate-based reparameterization of You~et~al.~\cite{zhonghui2019gate}, we associate a learnable scalar gate with each
convolutional output channel. For~a convolutional layer not followed by batch
normalization (BN), the~gate is applied directly to the convolution output.
The gates are initialized from the corresponding convolutional weights, which
are then divided by the initialized gate values to preserve the pretrained
mapping. Let \(\mathbf{X}\) denote the input feature map and \(\mathbf{W}\) denote the reparameterized convolutional weights. The~convolution output is \(\mathbf{X}\ast\mathbf{W}\in\mathbb{R}^{N\times C\times H\times W}\), where \(N\) is the batch size, \(C\) is the number of output channels, and~\(H\) and \(W\) are the height and width of the output feature map,
respectively. The~gated convolution output \(\mathbf{Y}\) is defined as 
\begin{equation}
    \mathbf{Y}
    =
    \boldsymbol{\phi}
    \odot
    \left(
    \mathbf{X}\ast\mathbf{W}
    \right),
\end{equation}
\textls[-15]{where \(\boldsymbol{\phi}\in\mathbb{R}^{C}\) is the learnable output-channel
gate vector, \(\ast\) denotes convolution, and~\(\odot\) denotes channel-wise
multiplication, with~\(\boldsymbol{\phi}\) broadcast over the batch and spatial
dimensions.}

When the convolution is followed by BN, the~gate is instead applied to the BN
output. To~preserve the pretrained mapping, \(\boldsymbol{\phi}\) is initialized
to the original BN scale parameter, the~BN shift parameter is divided by
\(\boldsymbol{\phi}\), and~the reparameterized BN scale parameter is set to
\(\mathbf{1}\) and held fixed during sparsity training. {Fixing the BN scale to \(\mathbf{1}\) transfers channel-wise scaling to the learnable gate, avoiding redundant scaling while preserving the pretrained mapping at initialization.} The gated BN output \(\mathbf{z}_{\mathrm{out}}\) is defined as
\begin{equation}
    \mathbf{z}_{\mathrm{out}}
    =
    \boldsymbol{\phi}
    \odot
    \left(
    \boldsymbol{\gamma}
    \odot
    \hat{\mathbf{z}}
    +
    \boldsymbol{\beta}
    \right),
\end{equation}
where
\(\hat{\mathbf{z}}\in\mathbb{R}^{N\times C\times H\times W}\) is the
normalized activation, and~\(\boldsymbol{\gamma}\in\mathbb{R}^{C}\) and
\(\boldsymbol{\beta}\in\mathbb{R}^{C}\) are the reparameterized BN scale and
shift parameters, respectively.

After sparsity training, the~gate magnitudes are used as channel-importance
scores. For~each retained channel, the~corresponding gate is folded into the
convolutional weights for layers without BN or into the BN affine parameters
for layers with BN, yielding standard layers without additional inference
operations.

\noindent\textbf{Gated Linear Layer.}
We further adapt the gate-based reparameterization to linear layers for
neuron-level pruning in the fully connected RoI head. Consider a linear layer
whose weight matrix and bias vector are denoted by
\(\mathbf{W}\in\mathbb{R}^{d_{\mathrm{out}}\times d_{\mathrm{in}}}\) and
\(\mathbf{b}\in\mathbb{R}^{d_{\mathrm{out}}}\), respectively, where
\(d_{\mathrm{in}}\) and \(d_{\mathrm{out}}\) denote the input and output
dimensions. We introduce a learnable gate vector
\(\boldsymbol{\psi}\in\mathbb{R}^{d_{\mathrm{out}}}\), with~each
\(\psi_i\) modulating the \(i\)-th output~neuron.

Each gate is initialized to the {root-mean-square (RMS)} magnitude of its corresponding
weight vector:
\begin{equation}
    \psi_i
    =
    \sqrt{
    \frac{\|\mathbf{w}_i\|_2^2}{d_{\mathrm{in}}}
    +
    \epsilon
    },
\end{equation}
where \(\mathbf{w}_i^{\top}\) denotes the \(i\)-th row of
\(\mathbf{W}\), and~\(\epsilon>0\) is a small constant for numerical
stability. {The RMS magnitude provides a dimension-normalized measure of the
pretrained neuron's weight scale. This gives the gate an informative starting
point for neuron importance estimation while reducing its dependence on the
input dimension, thereby improving comparability across linear layers of
different sizes.} Let \(b_i\) denote the \(i\)-th element of \(\mathbf{b}\).
To preserve the pretrained linear mapping upon gate insertion, we
reparameterize the weight vector and bias associated with the \(i\)-th output
neuron as
\begin{equation}
    \tilde{\mathbf{w}}_i
    =
    \frac{\mathbf{w}_i}{\psi_i},
    \qquad
    \tilde{b}_i
    =
    \frac{b_i}{\psi_i}.
\end{equation}

The 
gated output for the \(i\)-th neuron, denoted by \(y_i\), is given by
\begin{equation}
    y_i
    =
    \psi_i
    \left(
    \tilde{\mathbf{w}}_i^{\top}\mathbf{x}
    +
    \tilde{b}_i
    \right).
\end{equation}

Since
\(\psi_i\tilde{\mathbf{w}}_i=\mathbf{w}_i\) and
\(\psi_i\tilde{b}_i=b_i\), the~gated layer is initially equivalent to the
original pretrained linear~layer.

After sparsity training, the~gate magnitudes are used as neuron-importance
scores. For~each retained neuron, the~final weight vector
\(\mathbf{w}^{\star}_i\) and bias \(b^{\star}_i\) are obtained by folding the
corresponding gate into the reparameterized parameters:
\begin{equation}
    \mathbf{w}^{\star}_i
    =
    \psi_i\tilde{\mathbf{w}}_i,
    \qquad
    b^{\star}_i
    =
    \psi_i\tilde{b}_i.
\end{equation}

This yields a standard linear layer without additional inference~operations.

\subsubsection{Gate Sparsity Learning and Component-Wise~Pruning}
Starting from a pretrained detector, we train the gated model with an
\(\ell_1\) sparsity regularizer applied to all learnable gate vectors:
\begin{equation}
\mathcal{L}
=
\mathcal{L}_{\mathrm{det}}
+
\lambda_s
\sum_{\mathbf{g} \in \mathcal{G}}
\|\mathbf{g}\|_1,
\end{equation}
where \(\mathcal{L}_{\mathrm{det}}\) is the standard detection loss,
\(\mathcal{G}\) denotes the set of all learnable gate vectors, and~\(\mathbf{g}\in\mathcal{G}\) represents either a channel-wise gate vector
\(\boldsymbol{\phi}\) or a neuron-level gate vector
\(\boldsymbol{\psi}\). The~coefficient \(\lambda_s\) controls the strength of
the sparsity regularization. The~\(\ell_1\) penalty encourages unimportant
structures to have small gate magnitudes, which are subsequently used as
pruning~scores.

{After the gate parameters are learned through sparsity training, the~resulting gate magnitudes are used to perform component-wise structural
pruning.} Let
\(\mathcal{M}
=
\{\mathrm{backbone}, \mathrm{FPN}, \mathrm{RPN}, \mathrm{RoI\ head}\}\)
denote the set of detector components, and~let
\(\mathcal{G}_m\subseteq\mathcal{G}\) denote the set of gate vectors associated
with component \(m\in\mathcal{M}\). Rather than applying a single global
pruning threshold, we assign each component a target pruning ratio
\(\rho_m\) and determine the corresponding threshold \(\tau_m\) from the gate
magnitudes within that component. {The gate magnitudes of all gated layers within the same component are jointly ranked to determine a shared component-level threshold.} Specifically, \(\tau_m\) is chosen to remove a fraction \(\rho_m\) of the candidate channels or neurons with the smallest gate magnitudes. For~each gate vector \(\mathbf{g}\in\mathcal{G}_m\), the~initial retained index set
\(\mathcal{R}_m(\mathbf{g})\) is defined as
\begin{equation}
    \mathcal{R}_m(\mathbf{g})
    =
    \left\{
    i \mid |g_i| \geq \tau_m
    \right\},
\end{equation}
where \(g_i\) denotes the \(i\)-th element of \(\mathbf{g}\).

{To avoid excessively narrow layers and maintain hardware-friendly dimensions, the~number of retained channels or neurons in each layer is rounded up to the nearest multiple of 16, with~a minimum retained width of 16. When additional channels or neurons are required by this width-alignment constraint, those with the next largest gate magnitudes in the corresponding layer are retained.} The resulting pruning decisions are propagated across structurally coupled layers and branches, including adjacent convolutional and BN layers, skip connections, FPN branches, and~consecutive linear layers in the RoI head, to~maintain dimensional consistency. We then instantiate a compact detector from
the retained structures and fine-tune it to recover detection~accuracy.

\subsection{Knowledge~Distillation}
As shown in Figure \ref{fig:dtkd_framework}b, our distillation framework
trains a compressed student detector with ground-truth supervision and two
teacher branches in aligned rotated proposal spaces. We first describe the dual
teacher training scheme, then introduce rotated proposal alignment, and~finally
present the distillation~objective.

\subsubsection{Dual Teacher Knowledge~Distillation}
DTKD trains the compressed student using ground-truth supervision and two
teacher branches. The~homogeneous main teacher belongs to the same detector
family as the student and provides both classification and regression
supervision. The~heterogeneous auxiliary teacher is drawn from a different
detector family with distinct proposal-generation and RoI-representation mechanisms and provides additional classification~supervision.

{This design is motivated by the geometric diversity of SAR ships. Detectors
with different proposal-generation and RoI-representation mechanisms may exhibit
different strengths for ships with varying scales and aspect ratios. Therefore,
a single teacher may not provide uniformly strong guidance across all target
regimes.} The auxiliary teacher is selected to provide complementary
classification cues while maintaining sufficiently strong predictive
performance for reliable supervision. This design is further analyzed
in \mbox{Section \ref{sec:teacher_complementarity}}, {and the effect of the auxiliary
teacher choice is examined in Section \ref{sec:aux_teacher_analysis}.}

The auxiliary teacher is used only for classification distillation because its
classification logits and those of the student are defined over the same class
space and can be compared after proposal alignment. In~contrast, rotated box regression outputs are encoded as offsets relative to detector-specific proposals or reference boxes and depend on the corresponding coordinate transformations and box coders. Directly matching such heterogeneous offsets would, therefore, introduce inconsistent localization~supervision.

\subsubsection{Rotated Proposal~Alignment}

Prediction-level distillation in proposal-based detectors requires
teacher and student predictions to be evaluated on the same candidate
regions. Hence, we construct a rotated proposal-aligned space using
teacher-generated proposals. {At stage \(s\), let
\begin{equation}
\mathcal{P}_{t}^{(s)}
=
\{p_{t,i}^{(s)}\}_{i=1}^{N_t^{(s)}}
\end{equation}
denote the proposal set generated by teacher \(t\), where \(N_t^{(s)}\) is the
number of proposals and \(p_{t,i}^{(s)}\) represents the \(i\)-th oriented
candidate region. Both the teacher and the student are evaluated on each
proposal in \(\mathcal{P}_{t}^{(s)}\), thereby avoiding proposal mismatch and
providing proposal-wise aligned supervision at that stage. When \(s>0\), \(\mathcal{P}_{t}^{(s)}\) is obtained by refining the proposal set from the preceding stage. The~stage superscript \(s\) is omitted hereafter for notational simplicity.}

In SAR ship detection, not all aligned proposals play the same role. Large sea
areas and complex maritime backgrounds, including sea clutter, wakes,
shorelines, and~port structures, can produce proposals covering background or
cluttered regions. These proposals are useful for classification distillation
because they provide negative evidence for distinguishing ships from confusing
maritime backgrounds. However, they are unsuitable for regression distillation
because they do not have valid object-level regression~targets.

We therefore partition each aligned proposal set into foreground and background
subsets, denoted by \(\mathcal{P}_{t}^{\mathrm{fg}}\) and
\(\mathcal{P}_{t}^{\mathrm{bg}}\), respectively. A~proposal is assigned to the
foreground subset if its maximum rotated IoU with the ground-truth rotated boxes
is at least \(0.5\). All remaining proposals are assigned to the background
subset. The~proposal sets used for classification and regression distillation
are defined as
\begin{equation}
    \mathcal{P}_{t}^{\mathrm{cls}}
    =
    \mathcal{P}_{t}^{\mathrm{fg}}
    \cup
    \mathcal{P}_{t}^{\mathrm{bg}},
    \qquad
    \mathcal{P}_{t}^{\mathrm{reg}}
    =
    \mathcal{P}_{t}^{\mathrm{fg}}.
\end{equation}

Classification distillation is applied to
\(\mathcal{P}_{t}^{\mathrm{cls}}\), while regression distillation is restricted
to \(\mathcal{P}_{t}^{\mathrm{reg}}\). This preserves discriminative supervision
from challenging maritime background proposals while avoiding noisy
localization transfer from proposals without valid object-level regression
targets.

\subsubsection{Distillation~Objective}
The student is trained with the supervised detection loss and the proposed
dual-teacher distillation losses. Let \(\mathcal{L}_{\mathrm{det}}\) denote the
standard detection loss computed from ground-truth annotations. We use
\(t\in\{\mathrm{m},\mathrm{a}\}\) to denote the main and auxiliary teachers,
respectively. Both teachers provide classification distillation, while only the
main teacher provides regression~distillation.

For classification distillation, teacher and student logits are compared on the
aligned proposal set \(\mathcal{P}_{t}^{\mathrm{cls}}\). Given an aligned
proposal \(p_{t,i}\), let \(\mathbf{z}_{t,i}^{\mathrm{T}}\) and
\(\mathbf{z}_{t,i}^{\mathrm{S}}\) be the classification logits of the teacher
and the student on the same proposal, respectively. With~temperature \(\tau\), the~softened class distributions are
\begin{equation}
\mathbf{q}_{t,i}^{\mathrm{T}}
=
\mathrm{softmax}
\left(
\frac{\mathbf{z}_{t,i}^{\mathrm{T}}}{\tau}
\right)
\ \text{and}  \
\mathbf{q}_{t,i}^{\mathrm{S}}
=
\mathrm{softmax}
\left(
\frac{\mathbf{z}_{t,i}^{\mathrm{S}}}{\tau}
\right).
\end{equation}

The classification distillation loss from teacher \(t\) is defined using the {Kullback--Leibler (KL) divergence} as
\begin{equation}
\mathcal{L}_{t}^{\mathrm{cls}}
=
\frac{\tau^2}{|\mathcal{P}_{t}^{\mathrm{cls}}|}
\sum_{p_{t,i}\in \mathcal{P}_{t}^{\mathrm{cls}}}
\mathrm{KL}
\left(
\mathbf{q}_{t,i}^{\mathrm{T}}
\middle\|
\mathbf{q}_{t,i}^{\mathrm{S}}
\right),
\qquad
t \in \{\mathrm{m}, \mathrm{a}\}.
\end{equation}

Regression distillation is applied only to the foreground proposals of the main
teacher, \(\mathcal{P}_{\mathrm{m}}^{\mathrm{reg}}\). Let \(\mathbf{r}_{\mathrm{m},i}^{\mathrm{T}}\) and
\(\mathbf{r}_{\mathrm{m},i}^{\mathrm{S}}\) denote the encoded rotated-box
regression outputs of the main teacher and the student on the same proposal,
respectively.
The regression distillation \mbox{loss is}
\begin{equation}
\mathcal{L}_{\mathrm{m}}^{\mathrm{reg}}
=
\frac{1}{|\mathcal{P}_{\mathrm{m}}^{\mathrm{reg}}|}
\sum_{p_{\mathrm{m},i}\in \mathcal{P}_{\mathrm{m}}^{\mathrm{reg}}}
\mathrm{SmoothL1}
\left(
\mathbf{r}_{\mathrm{m},i}^{\mathrm{S}}
-
\mathbf{r}_{\mathrm{m},i}^{\mathrm{T}}
\right).
\end{equation}

The final training objective is
\begin{equation}
\mathcal{L}
=
\mathcal{L}_{\mathrm{det}}
+
\lambda_{\mathrm{m}}^{\mathrm{cls}}
\mathcal{L}_{\mathrm{m}}^{\mathrm{cls}}
+
\lambda_{\mathrm{m}}^{\mathrm{reg}}
\mathcal{L}_{\mathrm{m}}^{\mathrm{reg}}
+
\lambda_{\mathrm{a}}^{\mathrm{cls}}
\mathcal{L}_{\mathrm{a}}^{\mathrm{cls}},
\end{equation}
where \(\lambda_{\mathrm{m}}^{\mathrm{cls}}\),
\(\lambda_{\mathrm{m}}^{\mathrm{reg}}\), and~\(\lambda_{\mathrm{a}}^{\mathrm{cls}}\) balance main-teacher classification
distillation, main-teacher regression distillation, and~auxiliary-teacher
classification distillation, respectively.

\section{Results}
\label{sec:results}
This section presents the experimental evaluation of the proposed DTKDP framework.
We first describe the datasets, implementation details, and~evaluation metrics.
We then analyze the component-wise pruning scheme, teacher complementarity, and~the sensitivity to the distillation loss weights. Next, we compare the proposed
method with representative oriented object detectors and knowledge distillation
methods, followed by distillation ablation studies and auxiliary teacher
architecture analysis. Finally, we present qualitative results to further
verify the effectiveness of the proposed~design.

\subsection{Datasets}
Experiments are conducted on two public SAR ship detection datasets, namely SSDD~\cite{zhang2021ssdd} and RSDD-SAR~\cite{xu2022rsddsar}. SSDD is a widely used SAR ship detection dataset collected from multiple SAR sensors, including RadarSat-2, TerraSAR-X, and~Sentinel-1. It contains SAR images acquired with different polarization modes and spatial resolutions, covering both inshore and offshore scenes. In~this work, the~rotated bounding-box annotations of SSDD are adopted to evaluate oriented ship detection~performance.

RSDD-SAR is a rotated SAR ship detection dataset constructed from GF-3 and TerraSAR-X data. It consists of 7,000 image slices and 10,263 ship instances, covering multiple observation modes, polarization modes, and~spatial resolutions. The~dataset provides oriented bounding-box annotations for ship targets and includes ship instances with varying scales, aspect ratios, orientations, and~spatial distributions. These characteristics make it suitable for evaluating oriented SAR ship detectors in large-scale \mbox{experimental~settings.}

We follow the official data splits of both datasets. SSDD contains
928 training images and 232 test images, while RSDD-SAR contains 5000
training images and 2000 test images. {For model development, 15\% of
the official training set of each dataset is randomly held out as a
validation set for hyperparameter tuning. This validation split is fixed
and used consistently across all methods. Once the hyperparameters are
determined, the~training and validation subsets are merged for final
training, while the official test sets are reserved exclusively for final
evaluation. All reported results are obtained using the checkpoint from the
final training epoch.} The detailed characteristics of the two datasets are
summarized in~Table \ref{tab:datasets}.

\begin{table}[H]
\caption{Summary of the SAR ship detection datasets used in the~experiments.}
\label{tab:datasets}

\begin{tabularx}{\textwidth}{LLL}
\toprule
\textbf{Characteristic} & \textbf{SSDD} & \textbf{RSDD-SAR} \\
\midrule
Sensors
& RadarSat-2, TerraSAR-X, Sentinel-1
& Gaofen-3, TerraSAR-X \\

Polarization
& HH, HV, VH, VV
& HH, HV, VH, VV, DH, DV \\

Resolution (m)
& 1--15
& 2--20 \\

Scenes
& Inshore, offshore
& Inshore, offshore \\

Image size ($W \times H$)
& $214$--$668$ $\times$ $160$--$526$ px
& $512 \times 512$ px \\

Images (official train/test)
& 928/232
& 5000/2000 \\

Instances
& 2456
& 10,263 \\

Annotation
& HBB, OBB, PSeg
& OBB \\
\bottomrule
\end{tabularx}
\end{table}

\subsection{Implementation~Details}

All experiments are conducted on a single NVIDIA RTX 3090 GPU. The~detector
implementations are based on MMRotate~\cite{zhou2022mmrotate} and
MMYOLO~\cite{mmyolo2022}, while MMRazor~\cite{2021mmrazor} is used for the
knowledge distillation experiments. Baseline detectors are trained with
AdamW~\cite{AdamW}, following the corresponding MMRotate and MMYOLO
configurations. Models are trained for 150 epochs on SSDD with a batch size
of 1 and for 210 epochs on RSDD-SAR with a batch size of 4. The~initial
learning rates are set to \(3.75\times10^{-5}\) for two-stage detectors and
\(1.5\times10^{-4}\) for one-stage detectors, with~a step learning-rate
schedule following the corresponding baseline configurations. Linear warm-up
is applied for 1000 and 50 iterations, respectively. During~training, the~shorter image side is randomly sampled from 128 to 800 pixels, while the longer
side is limited to 1333 pixels. Images are normalized using ImageNet statistics
and augmented with random horizontal flips with a probability of 0.5. During~inference, all images are resized to \(512\times512\), and~no test-time
augmentation is used. {The \(512\times512\) resolution provides a common
evaluation scale for the two datasets while balancing target-detail
preservation and computational efficiency. The~confidence threshold is set
to 0.05 and the NMS IoU threshold to 0.1. For~proposal-based detectors, up~to 2000 highest scoring proposals are considered for NMS, and~at most 2000 proposals are retained per image. Unless~otherwise specified, experiments are run with a fixed
random seed of 42.} Sparsity training and knowledge distillation otherwise
follow the same training settings as the corresponding baseline detectors.
For DTKDP, Oriented R-CNN (R50) and RoI Transformer (R50) serve as both
teacher models and base detectors for~compression.

During sparsity training, gated versions of Oriented R-CNN (R50) and RoI
Transformer (R50) are further trained for 60 epochs on each dataset. The~sparsity coefficient \(\lambda_s\) is set to \(5\times10^{-5}\) for Oriented
R-CNN (R50) and \(5\times10^{-6}\) for RoI Transformer (R50). The~sparse
models are then pruned according to the learned gate values. The~target
backbone pruning ratios are set to 80\% for Oriented R-CNN (R50) and 60\% for
RoI Transformer (R50), while the remaining components use a target pruning
ratio of 50\%. The~effects of the sparsity coefficient and pruning strength
are further examined in~Section \ref{sec:component_pruning_analysis}. After~gated
channels and neurons are pruned, the~corresponding pruning decisions are
propagated through structurally coupled modules to maintain dimensional
consistency. Consequently, the~prescribed ratios specify the direct pruning
targets, while dependency propagation and width alignment may lead to
different overall parameter reductions. The~resulting compact detectors are
subsequently used as student models for knowledge~distillation.

The distillation loss weights are set according to the student detector. For~the Oriented R-CNN student, Oriented R-CNN (R50) serves as the main teacher,
while RoI Transformer (R50) serves as the auxiliary teacher. The~main-teacher
classification and regression distillation losses are assigned weights of 1.0
and 0.5, respectively, and~the auxiliary-teacher classification loss is
assigned a weight of 0.5. For~the RoI Transformer student, RoI Transformer
(R50) serves as the main teacher, while Oriented R-CNN (R50) serves as the
auxiliary teacher. Main-teacher distillation is performed at two cascade
stages. The~classification and regression distillation losses are weighted by
0.5 and 0.25 at stage 0, and~by 1.0 and 0.5 at stage 1, respectively. The~auxiliary teacher provides only the stage-1 classification distillation loss,
with a weight of 0.5. The~sensitivity of these distillation loss weights is further analyzed in Section \ref{sec:distill_weight_analysis}. {For the comparison with existing distillation
methods, each method is independently trained using three random seeds
(42, 43, and~44), and~the reported accuracy results are presented as
mean $\pm$ standard deviation over the three runs.}

\subsection{Evaluation~Metrics}

We evaluate each SAR ship detector in terms of detection accuracy and inference
efficiency. For~accuracy evaluation, average precision (AP) and mean average
precision (mAP) are computed using rotated intersection over union (IoU),
following standard object detection evaluation protocols~\cite{everingham2010pascal,lin2014coco}.
Given an IoU threshold \(t\), a~predicted rotated box is considered a true
positive if its rotated IoU with a ground-truth box is no smaller than \(t\).
Unmatched predictions are counted as false positives, whereas ground-truth
instances that are not matched by any prediction are counted as false negatives.
The corresponding precision and recall are defined as
\begin{equation}
\mathrm{P}_{t}
=
\frac{TP_{t}}{TP_{t}+FP_{t}}
\ \text{and} \
\mathrm{R}_{t}
=
\frac{TP_{t}}{TP_{t}+FN_{t}},
\end{equation}
where \(TP_{t}\), \(FP_{t}\), and~\(FN_{t}\) denote the numbers of true
positives, false positives, and~false negatives at threshold \(t\), respectively.
In the qualitative results, \(P@0.75\) and \(R@0.75\) denote precision and recall
computed at a rotated IoU threshold of~0.75.

For the same threshold \(t\), the~average precision \(\mathrm{AP}_{t}\) is
defined as the area under the precision--recall curve:
\begin{equation}
\mathrm{AP}_{t}
=
\int_{0}^{1} P_{t}(R)\,dR,
\end{equation}
where \(P_{t}(R)\) denotes precision as a function of recall \(R\) at threshold
\(t\). We report \(\mathrm{AP}_{50}\) and \(\mathrm{AP}_{75}\) to evaluate
detection accuracy under different localization~criteria.

To summarize performance over multiple IoU thresholds, we define
\begin{equation}
\mathrm{mAP}_{a:b}
=
\frac{1}{|\mathcal{T}_{a:b}|}
\sum_{t \in \mathcal{T}_{a:b}}
\mathrm{AP}_{t},
\qquad
\mathcal{T}_{a:b}
=
\{a, a+0.05, \ldots, b\}.
\end{equation}

Thus, \(\mathrm{mAP}_{50:95}\) denotes the mean AP over IoU thresholds from 0.50
to 0.95 with a step size of 0.05, following the COCO evaluation range. We
additionally report \(\mathrm{mAP}_{50:75}\) to characterize detection
performance over moderate-to-strict localization thresholds, {which are particularly relevant to small ships with limited pixel coverage or ships with large aspect ratios}.

For inference efficiency, we report the number of parameters, FLOPs, model size,
peak GPU memory, and~frames per second (FPS). These metrics quantify model
scale, computational cost, storage cost, inference memory demand, and~inference
speed, respectively. Peak GPU memory is measured as the maximum allocated GPU
memory during inference. All efficiency results are measured on the same
hardware platform under identical inference~settings.

\subsection{Component-Wise Pruning~Analysis}
\label{sec:component_pruning_analysis}
{This subsection analyzes the behavior and effectiveness of the proposed
component-wise pruning scheme. We first examine the sensitivity to the sparsity
coefficient $\lambda_s$. We then analyze the learned gate distributions and
the effect of pruning strength, followed by a matched-budget comparison of
backbone, global, and~component-wise pruning strategies. Finally, we summarize
the component-wise compression achieved by the adopted pruning configuration.}

\subsubsection{Sparsity Coefficient~Analysis}

Table \ref{tab:sparsity_sensitivity} {reports the detection performance and
gate sparsity of the gated models under different values of the sparsity
coefficient $\lambda_s$ before structural pruning. Across both datasets,
$S_{10^{-2}}$ increases consistently as $\lambda_s$ increases, indicating that
stronger sparsity regularization drives a larger fraction of gates toward zero.
Oriented R-CNN (R50) shows relatively high tolerance to stronger sparsity
regularization. At~$\lambda_s=5\times10^{-5}$, the~gated model achieves
mAP$_{50:75}$ values of 0.8314 and 0.7694, with~$S_{10^{-2}}$ values of
62.45\% and 74.20\% on SSDD and RSDD-SAR, respectively. Further increasing
$\lambda_s$ to $1\times10^{-4}$ leads to noticeable accuracy degradation.}

{RoI Transformer (R50) is more sensitive to stronger sparsity
regularization. At \linebreak $\lambda_s=5\times10^{-6}$, the~gated model achieves the
highest mAP$_{50:75}$ on both datasets, reaching 0.8398 on SSDD and 0.7660 on
RSDD-SAR. Larger coefficients further increase gate sparsity but lead to clear
accuracy degradation. These results indicate different sparsity sensitivities
between the two detector architectures. Accordingly, we adopt
$\lambda_s=5\times10^{-5}$ for Oriented R-CNN (R50) and
$\lambda_s=5\times10^{-6}$ for RoI Transformer (R50) in the subsequent
\mbox{pruning experiments}.}
\begin{table}[H]

\caption{Sensitivity 
analysis of the sparsity coefficient $\lambda_s$ on the validation sets of SSDD and RSDD-SAR.
$S_{10^{-2}}$ denotes the percentage of gates satisfying $|g_i|<10^{-2}$ after sparsity training.
Both detector architectures are evaluated over the same range of $\lambda_s$.
The adopted architecture-specific settings are highlighted in \textbf{bold}
($\uparrow$: higher values indicate greater sparsity).
}
\label{tab:sparsity_sensitivity}
\small

\begin{tabularx}{\textwidth}{CCCCC}
\toprule
&
\multicolumn{2}{c}{\textbf{Oriented R-CNN (R50)}} &
\multicolumn{2}{c}{\textbf{RoI Transformer (R50)}} \\
\cmidrule{2-5}

\textbf{$\boldsymbol{\lambda_s}$} &
\textbf{mAP$\boldsymbol{_{50:75}}$} &
\makecell{\textbf{$\boldsymbol{S_{10^{-2}}}$}\\\textbf{(\%)} $\boldsymbol{\uparrow}$} &
\textbf{mAP$\boldsymbol{_{50:75}}$} &
\makecell{\textbf{$\boldsymbol{S_{10^{-2}}}$}\\\textbf{(\%)} $\boldsymbol{\uparrow}$} \\
\midrule

\multicolumn{5}{l}{SSDD 
} \\
\addlinespace[1pt]

$2.5\times10^{-6}$
& 0.8409 & 47.72
& 0.8224 & 43.31 \\

$5\times10^{-6}$
& 0.8387 & 50.41
& \textbf{0.8398} & \textbf{47.07} \\

$1\times10^{-5}$
& 0.8321 & 53.78
& 0.7806 & 52.07 \\

$2.5\times10^{-5}$
& 0.8255 & 57.00
& 0.7671 & 58.63 \\

$5\times10^{-5}$
& \textbf{0.8314} & \textbf{62.45}
& 0.7383 & 61.32 \\

$1\times10^{-4}$
& 0.8083 & 65.95
& 0.7109 & 65.86 \\

\midrule

\multicolumn{5}{l}{RSDD} \\
\addlinespace[1pt]

$2.5\times10^{-6}$
& 0.7811 & 49.05
& 0.7503 & 52.74 \\

$5\times10^{-6}$
& 0.7769 & 56.28
& \textbf{0.7660} & \textbf{56.70} \\

$1\times10^{-5}$
& 0.7742 & 61.95
& 0.7471 & 59.79 \\

$2.5\times10^{-5}$
& 0.7723 & 67.33
& 0.7287 & 61.29 \\

$5\times10^{-5}$
& \textbf{0.7694} & \textbf{74.20}
& 0.6983 & 65.37 \\

$1\times10^{-4}$
& 0.7429 & 81.16
& 0.6627 & 69.91 \\

\bottomrule
\end{tabularx}
\end{table}
\unskip

\subsubsection{Gate Distribution and Pruning Strength~Analysis}

\begin{figure}[H]
\centering

\includegraphics[
    width=1.0\textwidth
]{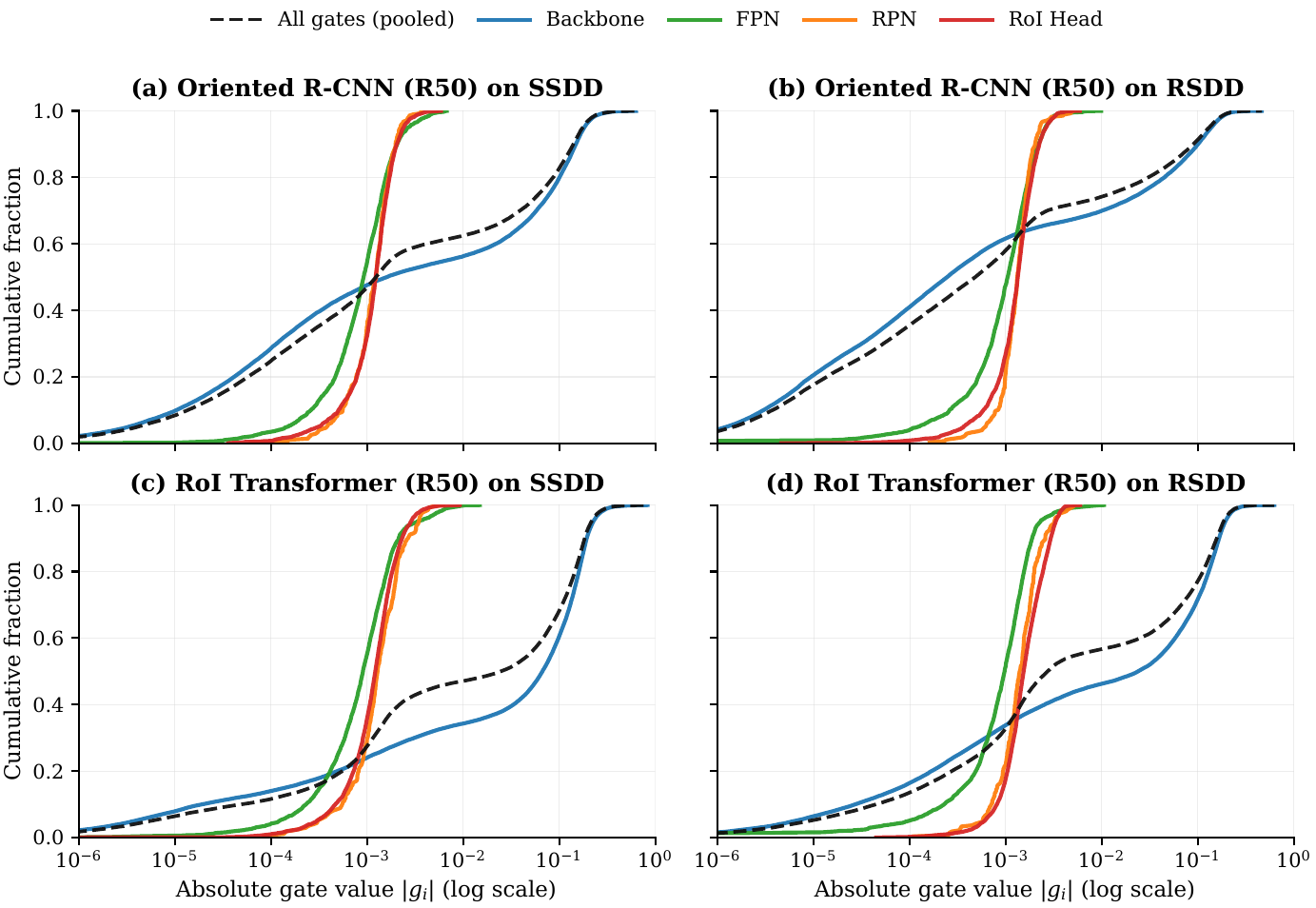}

\caption{
\textls[-15]{
Empirical cumulative distributions of absolute gate values after sparsity
training for Oriented R-CNN (R50) and RoI Transformer (R50) on SSDD and
RSDD-SAR. The~dashed curve pools gates from all components, while the solid
curves show the backbone, FPN, RPN, and~RoI head separately.
}
}

\label{fig:gate_distribution}
\end{figure}

{The learned gate distributions provide insight into the pruning behavior
of different detector components. As~shown in
Figure \ref{fig:gate_distribution}, the~FPN, RPN, and~RoI head exhibit relatively
concentrated gate distributions, whereas the backbone spans a broader range of
gate magnitudes. In~addition, the~backbone of Oriented R-CNN (R50) contains a
larger fraction of low-magnitude gates than that of RoI Transformer (R50) on
both datasets, suggesting greater structural redundancy and supporting a
stronger backbone pruning ratio. More importantly, a~common threshold
corresponds to different cumulative fractions across components because their
gate distributions are not aligned. This observation motivates determining
the pruning threshold separately within each detector component.}

{We examine the sensitivity to the target pruning strength in
Table \ref{tab:pruning_ratio_sensitivity}. For~Oriented R-CNN (R50), moving from
the Mild to the Medium setting reduces the parameter count by 51.03\% and
48.39\%, and~the FLOPs by 38.33\% and 35.11\%, on~SSDD and RSDD-SAR,
respectively, while the corresponding relative decreases in
mAP$_{50:75}$ are only 0.75\% and 0.92\%. A~similar trend is observed for
RoI Transformer (R50), where the Medium setting reduces the parameter count
by 38.26\% and 38.04\%, and~the FLOPs by 31.23\% and 28.29\%, on~SSDD and
RSDD-SAR, respectively, while the corresponding relative decreases in
mAP$_{50:75}$ are 1.94\% and 1.39\%. In~contrast, the~Strong setting provides
further compression but leads to more noticeable accuracy degradation on both
datasets. These results indicate that the Medium setting provides a favorable
balance between model complexity and detection accuracy and is therefore
adopted in the subsequent experiments.}

\vspace{-8pt}
\begin{table}[H]

\caption{Sensitivity 
analysis of the target pruning strength on the validation sets of
SSDD and RSDD-SAR. Each four-value pruning ratio is reported in the order of
backbone, FPN, RPN, and~RoI head. For~Oriented R-CNN (R50), the~Mild, Medium,
and Strong settings use ratios of 70/40/40/40, 80/50/50/50, and~90/60/60/60,
respectively. For~RoI Transformer (R50), the~corresponding ratios are
50/40/40/40, 60/50/50/50, and~70/60/60/60. The~adopted setting is highlighted
in \textbf{bold} ($\downarrow$: lower values are better).}
\label{tab:pruning_ratio_sensitivity}
\small

\begin{tabularx}{\textwidth}{lCCCCCC}
\toprule
&
\multicolumn{3}{c}{\textbf{SSDD}} &
\multicolumn{3}{c}{\textbf{RSDD}} \\
\cmidrule{2-7}

\textbf{Setting} &
\textbf{mAP$\boldsymbol{_{50:75}}$} &
\makecell{\textbf{Parameters}\\\textbf{(M)} $\boldsymbol{\downarrow}$} &
\makecell{\textbf{FLOPs}\\\textbf{(G)} $\boldsymbol{\downarrow}$} &
\textbf{mAP$\boldsymbol{_{50:75}}$} &
\makecell{\textbf{Parameters}\\\textbf{(M)} $\boldsymbol{\downarrow}$} &
\makecell{\textbf{FLOPs}\\\textbf{(G)} $\boldsymbol{\downarrow}$} \\
\midrule

\multicolumn{7}{l}{Oriented R-CNN (R50) 
} \\
\addlinespace[1pt]

Mild
& 0.8450 & 7.80 & 17.82
& 0.7811 & 6.84 & 17.29 \\

\textbf{Medium}
& \textbf{0.8387} & \textbf{3.82} & \textbf{10.99}
& \textbf{0.7739} & \textbf{3.53} & \textbf{11.22} \\

Strong
& 0.8008 & 1.61 & 7.79
& 0.7436 & 1.16 & 5.29 \\

\midrule

\multicolumn{7}{l}{RoI Transformer (R50)} \\
\addlinespace[1pt]

Mild
& 0.8470 & 11.24 & 19.31
& 0.7671 & 11.33 & 20.04 \\

\textbf{Medium}
& \textbf{0.8306} & \textbf{6.94} & \textbf{13.28}
& \textbf{0.7564} & \textbf{7.02} & \textbf{14.37} \\

Strong
& 0.8173 & 5.09 & 12.04
& 0.7342 & 5.09 & 11.13 \\

\bottomrule
\end{tabularx}
\end{table}
\unskip

\subsubsection{Comparison of Pruning~Strategies}

{We further compare Backbone Pruning, Global Pruning, and~Component-wise
Pruning under comparable parameter budgets on the validation sets, as~shown
in Table \ref{tab:pruning_strategy}. For~a fair comparison, the~pruning
configurations are adjusted separately so that the three strategies achieve
similar model sizes. Component-wise Pruning achieves the highest
mAP$_{50:75}$ for both detector architectures on both datasets. Compared with
Backbone Pruning, it yields relative mAP$_{50:75}$ improvements ranging from
0.83\% to 2.06\% across the four combinations of detector and dataset.
Compared with Global Pruning, the~corresponding improvements range from
0.89\% to 1.64\%. These results indicate that extending pruning beyond the
backbone to the full detection pipeline, including the FPN, RPN, and~RoI head,
provides a better trade-off between detection accuracy and model complexity
under comparable parameter budgets. They also demonstrate that using separate
pruning thresholds for individual components is more effective than applying
a single global threshold.}

\begin{table}[H]

\caption{Comparison 
of pruning strategies on the validation sets of SSDD and RSDD-SAR
under comparable parameter budgets. Backbone Pruning restricts gate-based
pruning to the backbone. Global Pruning uses a single pruning threshold for
the entire model, whereas Component-wise Pruning determines a separate
threshold for each detector component. The~highest mAP$_{50:75}$ for each
detector and dataset is highlighted in \textbf{bold}
($\downarrow$: lower values are better).
}
\label{tab:pruning_strategy}
\small

\begin{tabularx}{\textwidth}{lcccCcc}
\toprule
&
\multicolumn{3}{c}{\textbf{SSDD}} &
\multicolumn{3}{c}{\textbf{RSDD}} \\
\cmidrule{2-7}

\textbf{Pruning Strategy} &
\textbf{mAP$\boldsymbol{_{50:75}}$} &
\makecell{\textbf{Parameters}\\\textbf{(M)} $\boldsymbol{\downarrow}$} &
\makecell{\textbf{FLOPs}\\\textbf{(G)} $\boldsymbol{\downarrow}$} &
\textbf{mAP$\boldsymbol{_{50:75}}$} &
\makecell{\textbf{Parameters}\\\textbf{(M)} $\boldsymbol{\downarrow}$} &
\makecell{\textbf{FLOPs}\\\textbf{(G)} $\boldsymbol{\downarrow}$} \\
\midrule

\multicolumn{7}{l}{Oriented R-CNN (R50) 
} \\
\addlinespace[1pt]

Backbone Pruning
& 0.8326 & 23.34 & 44.36
& 0.7678 & 23.06 & 45.07 \\

Global Pruning
& 0.8361 & 22.99 & 42.34
& 0.7633 & 23.00 & 43.67 \\

Component-wise
& \textbf{0.8435} & 23.00 & 42.44
& \textbf{0.7742} & 23.26 & 42.43 \\

\midrule

\multicolumn{7}{l}{RoI Transformer (R50)} \\
\addlinespace[1pt]

Backbone Pruning
& 0.8357 & 34.99 & 44.37
& 0.7481 & 35.01 & 45.13 \\

Global Pruning
& 0.8352 & 35.78 & 46.01
& 0.7512 & 34.99 & 46.66 \\

Component-wise
& \textbf{0.8462} & 35.00 & 42.15
& \textbf{0.7635} & 34.31 & 41.42 \\

\bottomrule
\end{tabularx}
\end{table}
{Finally, Table \ref{tab:component_pruning} summarizes the component-wise
parameter reductions obtained with the adopted pruning configuration.
After the pruning configuration is selected on the validation set, the~training
and validation subsets are merged for final training.
Substantial compression is achieved across all detector components, including
the FPN, RPN, and~the parameter-intensive RoI head, which accounts for
approximately 33.6\% and 50.3\% of the parameters in the original Oriented
R-CNN (R50) and RoI Transformer (R50), respectively. The~adopted configuration
reduces the total parameter count by 90.3--91.8\% for Oriented R-CNN (R50) and
87.5--88.1\% for RoI Transformer (R50), while reducing the RoI head parameters
by 92.1--92.3\% and 96.8--97.5\%, respectively. These results confirm the
benefit of extending structured pruning beyond the backbone to the full
two-stage detection pipeline.}

\begin{table}[H]

\caption{Component-wise parameter counts before and after pruning.
Values are reported in millions, with relative reductions indicated by
$\downarrow$ in parentheses.}
\label{tab:component_pruning}
\small

\begin{tabularx}{\textwidth}{llCCC}
\toprule
\multirow{2.5}{*}{\textbf{Detector}} & \multirow{2.5}{*}{\textbf{Component}}
& \multirow{2.5}{*}{\textbf{Before Pruning}}
& \multicolumn{2}{c}{\textbf{After Pruning}} \\
\cmidrule{4-5}
& & & \textbf{SSDD} & \textbf{RSDD} \\
\midrule

\multirow{5}{*}{Oriented R-CNN (R50)}
& Backbone & 23.508 & 2.822 {\scriptsize($\downarrow$88.0\%)} & 2.076 {\scriptsize($\downarrow$91.2\%)} \\
& FPN      & 3.344  & 0.052 {\scriptsize($\downarrow$98.4\%)} & 0.099 {\scriptsize($\downarrow$97.0\%)} \\
& RPN      & 0.595  & 0.079 {\scriptsize($\downarrow$86.7\%)} & 0.116 {\scriptsize($\downarrow$80.5\%)} \\
& RoI Head & 13.903 & 1.070 {\scriptsize($\downarrow$92.3\%)} & 1.104 {\scriptsize($\downarrow$92.1\%)} \\
& Total 
& 41.350
& 4.023 {\scriptsize($\downarrow$90.3}\%)
& 3.395 {\scriptsize($\downarrow$91.8}\%) \\
\midrule

\multirow{5}{*}{RoI Transformer (R50)}
& Backbone & 23.508 & 5.936 {\scriptsize($\downarrow$74.7\%)} & 5.812 {\scriptsize($\downarrow$75.3\%)} \\
& FPN      & 3.344  & 0.027 {\scriptsize($\downarrow$99.2\%)} & 0.022 {\scriptsize($\downarrow$99.3\%)} \\
& RPN      & 0.594  & 0.041 {\scriptsize($\downarrow$93.1\%)} & 0.041 {\scriptsize($\downarrow$93.1\%)} \\
& RoI Head & 27.806 & 0.889 {\scriptsize($\downarrow$96.8\%)} & 0.682 {\scriptsize($\downarrow$97.5\%)} \\
& Total
& 55.252
& 6.893 {\scriptsize($\downarrow$87.5\%})
& 6.557 {\scriptsize($\downarrow$88.1\%}) \\
\bottomrule
\end{tabularx}
\end{table}
\unskip

\subsection{Teacher Complementarity~Analysis}
\label{sec:teacher_complementarity}

\begin{figure}[H]
\centering

\makebox[\textwidth][c]{%
    \includegraphics[
        width=1.25\textwidth
    ]{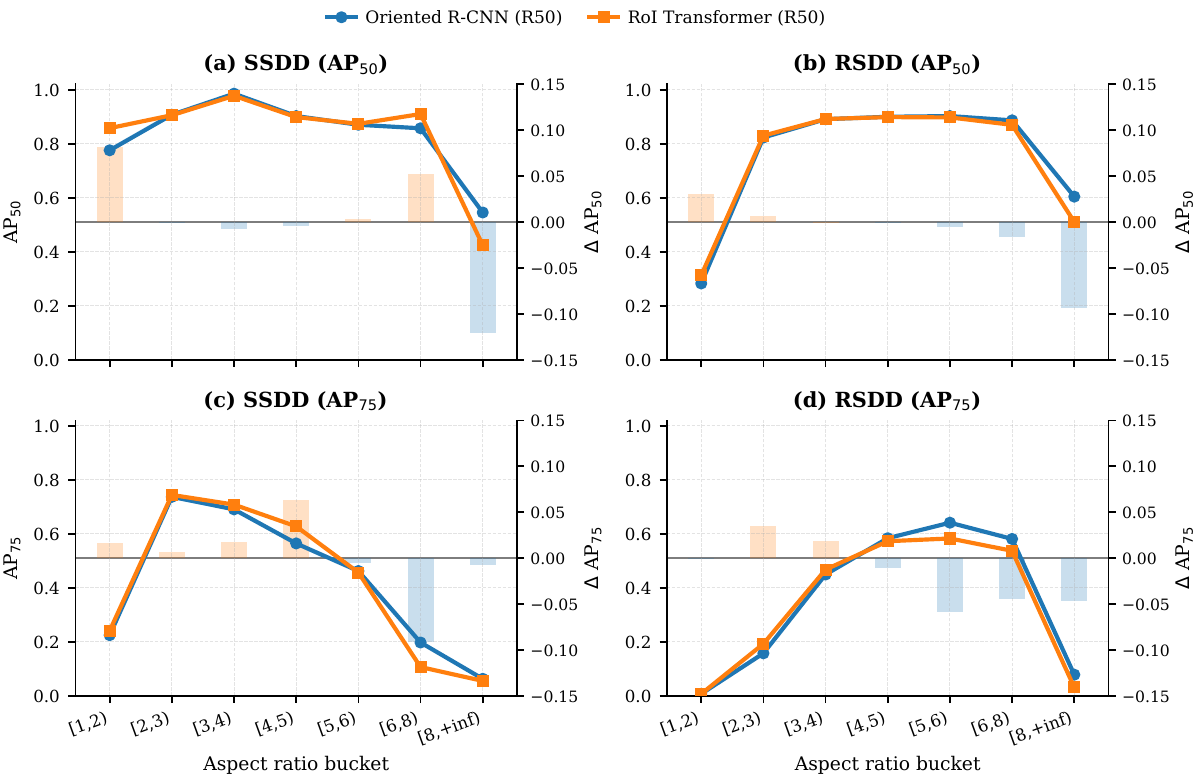}
}

\caption{
Teacher complementarity across ship aspect-ratio groups on SSDD and RSDD-SAR.
Blue and orange curves represent Oriented R-CNN (R50) and RoI Transformer (R50),
respectively. The translucent bars show
$\Delta\mathrm{AP}
=\mathrm{AP}(\mathrm{RoI\ Transformer})
-\mathrm{AP}(\mathrm{Oriented\ R\mbox{-}CNN})$.
Orange bars indicate positive differences, whereas blue bars indicate negative differences.
}

\label{fig:ar_motivation}
\end{figure}

To examine teacher complementarity, we evaluate the two teacher detectors across
ship aspect-ratio {and object-scale} groups on the SSDD and RSDD-SAR test sets.
As shown in~Figure \ref{fig:ar_motivation}, the~relative strengths of the two
teachers vary across aspect-ratio regimes {and localization criteria}.

On SSDD, RoI Transformer (R50) performs better in several aspect-ratio groups
at AP$_{50}$, including relative improvements of 10.57\% and 6.16\% in the
$[1,2)$ and $[6,8)$ buckets, respectively. In~contrast, Oriented R-CNN (R50)
outperforms RoI Transformer (R50) by 28.57\% in the most elongated
$[8,+\infty)$ bucket. {Under the stricter AP$_{75}$ criterion, RoI Transformer
(R50) performs better from the $[1,2)$ to $[4,5)$ buckets, with~an 11.17\%
relative improvement in the $[4,5)$ bucket. The~advantage then shifts toward
Oriented R-CNN (R50) for ships with larger aspect ratios. In~the $[6,8)$ bucket,
AP$_{75}$ increases from 0.1066 for RoI Transformer (R50) to 0.1976 for
Oriented R-CNN (R50).}

A similar trend is observed on RSDD-SAR. At~AP$_{50}$, RoI Transformer (R50)
leads in the lower aspect-ratio groups, achieving a 10.64\% relative improvement
in the $[1,2)$ bucket, whereas Oriented R-CNN (R50) leads from the $[4,5)$
bucket onward and achieves an 18.39\% relative improvement in the
$[8,+\infty)$ bucket. {At AP$_{75}$, RoI Transformer (R50) outperforms
Oriented R-CNN (R50) by 22.31\% in the $[2,3)$ bucket, while Oriented R-CNN
(R50) leads from the $[4,5)$ bucket onward, including relative improvements
of 10.03\% and 8.19\% in the $[5,6)$ and $[6,8)$ buckets, respectively.}

{The object-scale results in Table \ref{tab:scale_teacher_complementarity} further demonstrate this
complementarity. On~both datasets, RoI Transformer (R50) achieves higher
AP$_{50}$ and AP$_{75}$ for small ships, whereas Oriented R-CNN (R50) achieves
higher values for medium and large ships. On~SSDD, RoI Transformer (R50)
improves AP$_{50}$ and AP$_{75}$ for small ships by 10.51\% and 10.44\%,
respectively, while Oriented R-CNN (R50) improves AP$_{75}$ for medium ships
by 8.22\%. On~RSDD-SAR, RoI Transformer (R50) improves AP$_{75}$ for small
ships by 6.08\%, whereas Oriented R-CNN (R50) improves AP$_{75}$ by 4.71\%
and 15.18\% for medium and large \mbox{ships, respectively.}}

\textls[-15]{These results indicate that the two teacher architectures provide complementary
strengths rather than one consistently dominating the other. {RoI Transformer
(R50) performs better for compact and small ships in several settings, whereas
Oriented R-CNN (R50) shows stronger performance for elongated, medium, and~large ships, particularly under the stricter AP$_{75}$ criterion.} This
observation supports the use of a dual-teacher design in~DTKD.}

\begin{table}[H]
\footnotesize
\centering
\setlength{\tabcolsep}{4pt}
\renewcommand{\arraystretch}{1.08}

\caption{
Teacher detection performance on SSDD and RSDD-SAR across object-scale groups
defined by the normalized OBB scale
$s=\sqrt{wh/(WH)}$, where $w$ and $h$ denote the width and height of the ship
OBB, and $W$ and $H$ denote the width and height of the image, respectively.
Small, Medium, and Large correspond to
$s<t_{\mathrm{small}}$,
$t_{\mathrm{small}}\leq s<t_{\mathrm{large}}$, and
$s\geq t_{\mathrm{large}}$, respectively, where
$t_{\mathrm{small}}$ and $t_{\mathrm{large}}$ are the 33rd and 67th percentiles
of the scale distribution for each dataset.
$\Delta\mathrm{AP}
=\mathrm{AP}(\mathrm{RoI\ Transformer})
-\mathrm{AP}(\mathrm{Oriented\ R\mbox{-}CNN})$.
The better result for each scale group and metric is highlighted in
\textbf{bold}.
}
\label{tab:scale_teacher_complementarity}

\makebox[\textwidth][c]{%
\begin{tabularx}{1.06\textwidth}{
@{}
ll
*{6}{>{\centering\arraybackslash}X}
@{}
}

\toprule

\multirow{2}{*}[-6ex]{\textbf{Dataset}} &
\multirow{2}{*}[-6ex]{\textbf{Scale}} &
\multicolumn{3}{c}{\textbf{AP}$\boldsymbol{_{50}}$} &
\multicolumn{3}{c}{\textbf{AP}$\boldsymbol{_{75}}$} \\

\cmidrule(lr){3-5}
\cmidrule(lr){6-8}

& &
\makecell{\textbf{Oriented}\\\textbf{R-CNN (R50)}} &
\makecell{\textbf{RoI Transformer}\\\textbf{(R50)}} &
$\boldsymbol{\Delta}$\textbf{AP} &
\makecell{\textbf{Oriented}\\\textbf{R-CNN (R50)}} &
\makecell{\textbf{RoI Transformer}\\\textbf{(R50)}} &
$\boldsymbol{\Delta}$\textbf{AP} \\

\midrule

\multirow{3}{*}{SSDD}
& Small
& 0.7860
& \textbf{0.8686}
& +0.0826
& 0.2260
& \textbf{0.2496}
& +0.0235 \\

& Medium
& \textbf{0.9048}
& 0.9021
& -0.0027
& \textbf{0.6396}
& 0.5910
& -0.0486 \\

& Large
& \textbf{0.9081}
& 0.9070
& -0.0011
& \textbf{0.7720}
& 0.7700
& -0.0020 \\

\midrule

\multirow{3}{*}{RSDD}
& Small
& 0.7748
& \textbf{0.7825}
& +0.0077
& 0.2368
& \textbf{0.2512}
& +0.0144 \\

& Medium
& \textbf{0.9014}
& 0.8995
& -0.0019
& \textbf{0.5400}
& 0.5157
& -0.0243 \\

& Large
& \textbf{0.9068}
& 0.9032
& -0.0036
& \textbf{0.7081}
& 0.6148
& -0.0933 \\

\bottomrule

\end{tabularx}%
}

\end{table}
\unskip

\subsection{Distillation Loss Weight~Analysis}
\label{sec:distill_weight_analysis}

{We analyze the sensitivity of the distillation loss weights on the SSDD
validation set for both student architectures. For~each comparison, one group
of loss weights is varied while the remaining weights are kept fixed. A~zero-weight configuration is additionally included as the no-distillation
baseline. For~RoI Transformer-slim, each pair of main-teacher classification
or regression weights corresponds to cascade stages 0 and 1, respectively.
The distillation temperature is fixed at $\tau=1.0$. The~results are reported
in Tables \ref{tab:orcnn_distill_weight_sensitivity} and
\ref{tab:roi_distill_weight_sensitivity}.}

{For Oriented R-CNN-slim, the~configuration
$\lambda_{m}^{\mathrm{cls}}=1.0$,
$\lambda_{m}^{\mathrm{reg}}=0.5$, and~$\lambda_{a}^{\mathrm{cls}}=0.5$ achieves the highest AP$_{75}$,
mAP$_{50:75}$, and~mAP$_{50:95}$. Relative to this configuration,
increasing $\lambda_{m}^{\mathrm{cls}}$ to 2.0 leads to relative decreases
of 1.86\% and 3.19\% in mAP$_{50:75}$ and mAP$_{50:95}$,
respectively, whereas reducing it to 0.5 causes only minor changes in the two
mAP metrics but lowers AP$_{75}$. Reducing
$\lambda_{m}^{\mathrm{reg}}$ to 0.25 leads to relative decreases of 2.16\%
and 2.28\% in mAP$_{50:75}$ and mAP$_{50:95}$, respectively, while
increasing it to 1.0 mainly affects mAP$_{50:95}$. In~comparison, the~auxiliary classification weight shows milder sensitivity, with~both
$\lambda_{a}^{\mathrm{cls}}=0.25$ and 1.0 retaining competitive
performance.}

{A similar pattern is observed for RoI Transformer-slim. The~configuration
with main-teacher classification weights $(0.50,1.00)$, regression weights
$(0.25,0.50)$, and~auxiliary classification weight 0.50 achieves the highest
AP$_{50}$, mAP$_{50:75}$, and~mAP$_{50:95}$, while its AP$_{75}$ is close
to the highest value among the evaluated settings. Relative to this
configuration, increasing the classification weights to $(1.00,2.00)$ leads
to relative decreases of 0.87\% and 1.64\% in mAP$_{50:75}$ and
mAP$_{50:95}$, respectively. Reducing the regression weights to
$(0.125,0.25)$ leads to larger relative decreases of 1.37\% and 2.25\%,
whereas increasing them to $(0.50,1.00)$ results in only minor overall
changes. The~auxiliary classification weight again shows comparatively mild
sensitivity, with~both 0.25 and 1.0 producing performance close to that
obtained with 0.50.}

{Overall, the~results indicate that the main-teacher classification and
regression losses require appropriate balancing, whereas the auxiliary
classification loss is comparatively less sensitive over the evaluated range.
For both student architectures, all evaluated nonzero distillation
configurations outperform the corresponding no-distillation baseline in
AP$_{75}$, mAP$_{50:75}$, and~mAP$_{50:95}$. This indicates that the benefit
of dual-teacher distillation is maintained across a range of loss-weight
settings rather than depending on a single specific configuration. Based on
this analysis, we set
$\lambda_{m}^{\mathrm{cls}}=1.0$,
$\lambda_{m}^{\mathrm{reg}}=0.5$, and~$\lambda_{a}^{\mathrm{cls}}=0.5$ for Oriented R-CNN-slim, and~$\lambda_{m}^{\mathrm{cls}}=(0.50,1.00)$,
$\lambda_{m}^{\mathrm{reg}}=(0.25,0.50)$, and~$\lambda_{a}^{\mathrm{cls}}=0.50$ for RoI Transformer-slim in the
subsequent experiments.}

\begin{table}[H]
\small
\caption{
Sensitivity 
analysis of the distillation loss weights for Oriented R-CNN-slim
on the SSDD validation set. The~best result for each metric is highlighted in
\textbf{bold}.
}
\label{tab:orcnn_distill_weight_sensitivity}

\begin{tabularx}{\textwidth}{CCCCCCC}
\toprule
$\boldsymbol{\lambda_{m}^{\mathrm{cls}}}$ &
$\boldsymbol{\lambda_{m}^{\mathrm{reg}}}$ &
$\boldsymbol{\lambda_{a}^{\mathrm{cls}}}$ &
\textbf{AP}$\boldsymbol{_{50}}$ &
\textbf{AP}$\boldsymbol{_{75}}$ &
\textbf{mAP}$\boldsymbol{_{50:75}}$ &
\textbf{mAP}$\boldsymbol{_{50:95}}$ \\
\midrule

0 & 0 & 0
& 0.9069 & 0.6739 & 0.8310 & 0.5815 \\

0.50 & 0.50 & 0.50
& \textbf{0.9082} & 0.6880 & 0.8491 & 0.6034 \\

1.00 & 0.50 & 0.50
& 0.9079 & \textbf{0.6952} & \textbf{0.8500} & \textbf{0.6053} \\

2.00 & 0.50 & 0.50
& 0.9078 & 0.6934 & 0.8342 & 0.5860 \\

1.00 & 0.25 & 0.50
& 0.9074 & 0.6840 & 0.8316 & 0.5915 \\

1.00 & 1.00 & 0.50
& 0.9079 & 0.6871 & 0.8475 & 0.5894 \\

1.00 & 0.50 & 0.25
& 0.9076 & 0.6851 & 0.8457 & 0.5935 \\

1.00 & 0.50 & 1.00
& 0.9081 & 0.6909 & 0.8492 & 0.6050 \\

\bottomrule
\end{tabularx}
\end{table}
\unskip

\begin{table}[H]
\small
\caption{Sensitivity 
analysis of the distillation loss weights for RoI Transformer-slim
on the SSDD validation set. Each paired main-teacher weight is reported in the
order of cascade stages 0 and 1. The~best result for each metric is highlighted
in \textbf{bold}.
}
\label{tab:roi_distill_weight_sensitivity}

\begin{tabularx}{\textwidth}{CCCCCCC}
\toprule
$\boldsymbol{\lambda_{m}^{\mathrm{cls}}}$ &
$\boldsymbol{\lambda_{m}^{\mathrm{reg}}}$ &
$\boldsymbol{\lambda_{a}^{\mathrm{cls}}}$ &
\textbf{AP}$\boldsymbol{_{50}}$ &
\textbf{AP}$\boldsymbol{_{75}}$ &
\textbf{mAP}$\boldsymbol{_{50:75}}$ &
\textbf{mAP}$\boldsymbol{_{50:95}}$ \\
\midrule

$(0, 0)$ & $(0, 0)$ & 0
& 0.9070 & 0.6613 & 0.8243 & 0.5685 \\

$(0.25, 0.50)$ & $(0.25, 0.50)$ & 0.50
& 0.9074 & 0.6691 & 0.8434 & 0.5876 \\

$(0.50, 1.00)$ & $(0.25, 0.50)$ & 0.50
& \textbf{0.9079} & 0.6748 & \textbf{0.8461} & \textbf{0.5921} \\

$(1.00, 2.00)$ & $(0.25, 0.50)$ & 0.50
& 0.9072 & 0.6662 & 0.8387 & 0.5824 \\

$(0.50, 1.00)$ & $(0.125, 0.25)$ & 0.50
& 0.9069 & 0.6627 & 0.8345 & 0.5788 \\

$(0.50, 1.00)$ & $(0.50, 1.00)$ & 0.50
& 0.9077 & \textbf{0.6756} & 0.8449 & 0.5905 \\

$(0.50, 1.00)$ & $(0.25, 0.50)$ & 0.25
& 0.9073 & 0.6698 & 0.8428 & 0.5879 \\

$(0.50, 1.00)$ & $(0.25, 0.50)$ & 1.00
& 0.9078 & 0.6737 & 0.8455 & 0.5916 \\

\bottomrule
\end{tabularx}
\end{table}
\unskip

\subsection{Comparison with State of the~Art}

This subsection presents a comprehensive evaluation of the proposed method on two~representative SAR ship detection benchmarks, SSDD and RSDD-SAR. We compare the proposed compressed detectors with representative oriented object detectors from both two-stage and one-stage paradigms, including RTMDet variants that serve as strong, real-time efficient baselines, to~assess the trade-off between detection accuracy and computational efficiency. In~addition, the~effectiveness of the proposed distillation strategy is evaluated against representative knowledge distillation methods under identical student architectures and training settings. Specifically, we examine whether the compressed models can retain the localization capability of their full-scale teacher networks while achieving substantial reductions in parameters and FLOPs. We further investigate efficiency gains in terms of model size, memory consumption, and~inference speed, which are critical for deployment in resource-constrained scenarios. Overall, this subsection demonstrates that the proposed framework achieves a favorable accuracy--efficiency trade-off while maintaining competitive localization capability, and~that the proposed distillation mechanism consistently improves performance over existing feature-based distillation approaches for oriented SAR ship~detection.

\subsubsection{Comparison with Oriented Object~Detectors} 

We compare the proposed compressed detectors with representative oriented object
detectors on SSDD and RSDD-SAR, covering both two-stage and one-stage paradigms,
to evaluate the trade-off between detection accuracy and computational
efficiency. As~shown in~Tables \ref{tab:ssdd_comparison}
and~\ref{tab:rsdd_comparison}, the~compared methods include three groups:
(i) full two-stage detectors, (ii) one-stage detectors, including real-time
efficient RTMDet variants, and~(iii) the proposed slim two-stage detectors obtained through component-wise pruning and knowledge distillation, namely Oriented R-CNN-slim and RoI~Transformer-slim.

\noindent\textbf{Comparison with two-stage detectors. 
}
On SSDD, Oriented R-CNN-slim slightly outperforms its full-scale Oriented R-CNN (R50) counterpart across all four accuracy metrics, with~gains of 0.08\%, 0.65\%, 0.24\%, and~0.08\% in AP$_{50}$, AP$_{75}$, mAP$_{50:75}$, and~mAP$_{50:95}$, respectively. These improvements are achieved while reducing the parameter count by 90.3\%, FLOPs by 78.1\%, and~peak GPU memory by 54.6\%. The~inference speed also increases by {19.0\%}, from~{27.17} to {32.34} FPS. The~compressed RoI Transformer exhibits a different but still favorable trade-off. Relative to RoI Transformer (R50), RoI Transformer-slim improves AP$_{50}$ by 0.08\%, whereas AP$_{75}$, mAP$_{50:75}$, and~mAP$_{50:95}$ decrease by 2.38\%, 0.51\%, and~1.88\%, respectively. In~exchange, its parameter count, FLOPs, and~peak GPU memory are reduced by 87.5\%, 75.6\%, and~65.7\%, while its inference speed increases by {30.6\%}, from~{24.28} to {31.72} FPS. The~SSDD results therefore demonstrate lossless compression of Oriented R-CNN (R50), while RoI Transformer-slim retains competitive accuracy with substantial efficiency~gains.

On RSDD-SAR, both compressed models also achieve substantial efficiency gains while keeping the loss of detection accuracy limited. Compared with Oriented R-CNN (R50), Oriented R-CNN-slim records decreases of 0.17\%, 0.51\%, 1.70\%, and~1.18\% in AP$_{50}$, AP$_{75}$, mAP$_{50:75}$, and~mAP$_{50:95}$, respectively. Nevertheless, it reduces the parameter count by 91.8\%, FLOPs by 79.9\%, and~peak GPU memory by 56.0\%, while improving the inference speed by {13.8\%}, from~{30.61} to {34.82} FPS. RoI Transformer-slim follows a similar pattern. Relative to RoI Transformer (R50), AP$_{50}$, AP$_{75}$, mAP$_{50:75}$, and~mAP$_{50:95}$ decrease by 0.49\%, 2.11\%, 1.05\%, and~1.09\%, respectively. At~the same time, the~parameter count, FLOPs, and~peak GPU memory are reduced by 88.1\%, 77.2\%, and~67.5\%, respectively, while the inference speed increases by {31.1\%}, from~{23.71} to {31.08} FPS. Across the two detector families, the~largest accuracy reduction is only 2.11\%, whereas the parameter count and FLOPs are reduced by at least 88.1\% and 77.2\%, respectively. These results show that DTKDP substantially lowers the computational, storage, and~memory demands of two-stage oriented detectors without materially compromising their detection~performance.

\noindent\textbf{Comparison with one-stage detectors.}
Among the evaluated one-stage detectors, RTMDet-tiny serves as the primary lightweight baseline because its parameter count is comparable to those of the proposed models. On~SSDD, Oriented R-CNN-slim improves AP$_{50}$, AP$_{75}$, mAP$_{50:75}$, and~mAP$_{50:95}$ over RTMDet-tiny by 0.55\%, 27.55\%, 8.64\%, and~14.39\%, respectively. RoI Transformer-slim achieves corresponding gains of 0.52\%, 23.59\%, 8.19\%, and~12.47\%. The~particularly large improvements in AP$_{75}$ highlight the benefit of proposal generation and RoI-level refinement for strict rotated localization. Oriented R-CNN-slim also uses 17.5\% fewer parameters than RTMDet-tiny. Although~RTMDet-tiny requires fewer FLOPs and less peak GPU memory and runs faster, Oriented R-CNN-slim and RoI Transformer-slim still achieve {32.34} and {31.72} FPS, respectively. The~proposed models therefore deliver substantially stronger localization accuracy while maintaining competitive inference throughput under the tested RTX 3090 configuration. The~same pattern holds on RSDD-SAR. Compared with RTMDet-tiny, Oriented R-CNN-slim improves AP$_{50}$, AP$_{75}$, mAP$_{50:75}$, and~mAP$_{50:95}$ by 1.43\%, 20.17\%, 7.40\%, and~7.70\%, respectively, while using 30.2\% fewer parameters. RoI Transformer-slim provides corresponding gains of 1.08\%, 15.40\%, 5.91\%, and~5.14\%. RTMDet-tiny remains more efficient in FLOPs, peak GPU memory, and~inference speed, whereas the proposed models retain throughputs of {34.82} and {31.08} FPS. Taken together, these results show that the compressed two-stage detectors provide a favorable trade-off when strict rotated localization is prioritized over minimum inference~cost.

Overall, DTKDP substantially compresses two-stage detectors while preserving their localization performance. {Compared with the lightweight one-stage baseline, the~compressed models
achieve markedly higher accuracy under strict rotated IoU thresholds, at~the
cost of higher FLOPs, peak GPU memory, and~lower FPS than RTMDet-tiny. On~the
tested RTX 3090 platform, both compressed detectors achieve above 30 FPS on
both datasets, indicating real-time inference on this hardware
platform~\cite{wang2025rtdetrv3}.}

\begin{table}[H]
\footnotesize
\centering
\setlength{\tabcolsep}{3.5pt}
\renewcommand{\arraystretch}{1.05}

\caption{
Comparison with representative oriented object detectors on the SSDD dataset.
Baseline detectors are grouped by detection paradigm and sorted by the number
of parameters in descending order. $^\dagger$ denotes our compressed models
obtained by pruning and knowledge distillation.
Arrows indicate the preferred direction
($\uparrow$: higher is better; $\downarrow$: lower is better).
The best results are shown in \textbf{bold}, and the second-best results are
\underline{underlined}. FPS is reported as mean $\pm$ standard deviation over
three inference runs.
}
\label{tab:ssdd_comparison}

\makebox[\textwidth][c]{%
\begin{tabularx}{1.12\textwidth}{@{}Xccccccccc@{}}

\toprule

\textbf{Model} &
\textbf{AP$\boldsymbol{_{50}}$} &
\textbf{AP$\boldsymbol{_{75}}$} &
\textbf{mAP$\boldsymbol{_{50:75}}$} &
\textbf{mAP$\boldsymbol{_{50:95}}$} &
\makecell{\textbf{Parameters}\\\textbf{(M)} $\boldsymbol{\downarrow}$} &
\makecell{\textbf{FLOPs}\\\textbf{(G)} $\boldsymbol{\downarrow}$} &
\makecell{\textbf{Model Size}\\\textbf{(MB)} $\boldsymbol{\downarrow}$} &
\makecell{\textbf{Memory}\\\textbf{(MB)} $\boldsymbol{\downarrow}$} &
\textbf{FPS} $\boldsymbol{\uparrow}$ \\

\midrule


\multicolumn{10}{@{}l@{}}{\textbf{Two-Stage Detectors}} \\
\addlinespace[1pt]

\hspace*{0.8em}RoI Transformer (R50)~\cite{Ding_2019_CVPR}
& 0.9069
& 0.6892
& \textbf{0.8496}
& \textbf{0.6020}
& 55.25
& 56.91
& 210.97
& 528.47
& $24.28 \pm 0.55$ \\

\hspace*{0.8em}RoI Transformer (R18)~\cite{Ding_2019_CVPR}
& 0.9065
& 0.6722
& 0.8443
& 0.5919
& 42.18
& 43.39
& 160.95
& 477.31
& $29.78 \pm 0.32$ \\

\hspace*{0.8em}Oriented R-CNN (R50)~\cite{Xie_2021_ICCV}
& 0.9072
& \underline{0.6899}
& 0.8468
& 0.6003
& 41.35
& 53.30
& 157.94
& 375.48
& $27.17 \pm 0.49$ \\

\hspace*{0.8em}R-Faster R-CNN (R50)~\cite{ren2017faster}
& 0.9052
& 0.6444
& 0.8178
& 0.5507
& 41.35
& 53.10
& 157.94
& 475.11
& $31.63 \pm 0.31$ \\

\hspace*{0.8em}Oriented R-CNN (R18)~\cite{Xie_2021_ICCV}
& 0.9066
& 0.6769
& 0.8312
& 0.5916
& 28.28
& 39.78
& 107.92
& 322.34
& $31.12 \pm 0.35$ \\

\hspace*{0.8em}R-Faster R-CNN (R18)~\cite{ren2017faster}
& 0.9049
& 0.6326
& 0.8141
& 0.5481
& 28.28
& 39.58
& 107.92
& 422.16
& $36.05 \pm 0.54$ \\

\midrule


\multicolumn{10}{@{}l@{}}{\textbf{One-Stage Detectors}} \\
\addlinespace[1pt]

\hspace*{0.8em}S$^2$ANet (R50)~\cite{han2022s2anet}
& 0.9075
& 0.5739
& 0.7977
& 0.5424
& 38.76
& 49.05
& 148.06
& 317.31
& $22.26 \pm 0.24$ \\

\hspace*{0.8em}R-RetinaNet (R50)~\cite{lin2017focal}
& 0.9009
& 0.5639
& 0.7774
& 0.5360
& 36.35
& 52.39
& 138.87
& 308.11
& $31.80 \pm 0.29$ \\

\hspace*{0.8em}R-FCOS (R50)~\cite{fcosr}
& 0.9044
& 0.6498
& 0.8190
& 0.5713
& 32.12
& 51.55
& 122.71
& 291.96
& $33.12 \pm 0.41$ \\

\hspace*{0.8em}RTMDet-m~\cite{lyu2022rtmdet}
& \textbf{0.9085}
& 0.5785
& 0.7994
& 0.5490
& 24.66
& 25.01
& 104.40
& 140.95
& $37.18 \pm 0.67$ \\

\hspace*{0.8em}S$^2$ANet (R18)~\cite{han2022s2anet}
& 0.9061
& 0.5687
& 0.7944
& 0.5396
& 22.20
& 36.11
& 84.73
& 229.30
& $27.56 \pm 0.69$ \\

\hspace*{0.8em}R-RetinaNet (R18)~\cite{lin2017focal}
& 0.9008
& 0.5636
& 0.7701
& 0.5328
& 19.79
& 39.46
& 75.54
& 220.36
& $37.76 \pm 0.98$ \\

\hspace*{0.8em}R-FCOS (R18)~\cite{fcosr}
& 0.9028
& 0.6490
& 0.8160
& 0.5692
& 19.10
& 38.84
& 72.88
& 217.45
& $39.48 \pm 1.05$ \\

\hspace*{0.8em}RTMDet-s~\cite{lyu2022rtmdet}
& 0.9061
& 0.5684
& 0.7822
& 0.5330
& 8.86
& \underline{9.44}
& 38.38
& \underline{65.00}
& $\underline{42.92 \pm 0.76}$ \\

\hspace*{0.8em}RTMDet-tiny~\cite{lyu2022rtmdet}
& 0.9029
& 0.5444
& 0.7813
& 0.5252
& \underline{4.87}
& \textbf{5.14}
& \underline{21.19}
& \textbf{42.81}
& $\boldsymbol{45.24 \pm 0.97}$ \\

\midrule


\multicolumn{10}{@{}l@{}}{\textbf{Ours}} \\
\addlinespace[1pt]

\hspace*{0.8em}\textbf{RoI Transformer-slim $^\dagger$}
& 0.9076
& 0.6728
& 0.8453
& 0.5907
& 6.89
& 13.89
& 26.37
& 181.33
& $31.72 \pm 0.34$ \\

\hspace*{0.8em}\textbf{Oriented R-CNN-slim $^\dagger$}
& \underline{0.9079}
& \textbf{0.6944}
& \underline{0.8488}
& \underline{0.6008}
& \textbf{4.02}
& 11.69
& \textbf{15.39}
& 170.36
& $32.34 \pm 0.37$ \\

\bottomrule

\end{tabularx}%
}

\end{table}

\subsubsection{Comparison with Distillation~Methods}

We further compare DTKD with representative knowledge distillation methods,
including CWD~\cite{shu2021cwd}, PKD~\cite{cao2022pkd}, and~MGD~\cite{yang2022mgd}, using the same compressed student architectures and
training settings. As~shown in~Table \ref{tab:kd_comparison_ssdd_rsdd}, DTKD
achieves the best {mean performance} across all four accuracy metrics for
both student detectors on SSDD and RSDD-SAR. For~each metric, the~relative
improvement is calculated against the best {mean performance} among the
competing~methods.

Specifically, for~Oriented R-CNN-slim, CWD is the strongest competing method
across all four metrics on both datasets. On~SSDD, DTKD improves AP$_{50}$,
AP$_{75}$, mAP$_{50:75}$, and~mAP$_{50:95}$ over CWD by
{0.08\%, 3.04\%, 0.52\%, and~1.95\%}, respectively.
The corresponding improvements on RSDD-SAR are
{0.09\%, 0.81\%, 0.16\%, and~0.84\%}, respectively.
The relatively larger gains in AP$_{75}$ and mAP$_{50:95}$ indicate that the
benefits of DTKD are more evident under stricter rotated IoU~criteria.

\begin{table}[H]
\footnotesize
\centering
\setlength{\tabcolsep}{3.5pt}
\renewcommand{\arraystretch}{1.05}

\caption{
Comparison with representative oriented object detectors on the RSDD-SAR dataset.
Baseline detectors are grouped by detection paradigm and sorted by the number
of parameters in descending order.
$^\dagger$ denotes our compressed models obtained by pruning and knowledge distillation.
Arrows indicate the preferred direction
($\uparrow$: higher is better; $\downarrow$: lower is better).
The best results are shown in \textbf{bold}, and the second-best results are
\underline{underlined}.
FPS is reported as mean $\pm$ standard deviation over three inference runs.
}
\label{tab:rsdd_comparison}

\makebox[\textwidth][c]{%
\begin{tabularx}{1.12\textwidth}{@{}Xccccccccc@{}}

\toprule

\textbf{Model} &
\textbf{AP$\boldsymbol{_{50}}$} &
\textbf{AP$\boldsymbol{_{75}}$} &
\textbf{mAP$\boldsymbol{_{50:75}}$} &
\textbf{mAP$\boldsymbol{_{50:95}}$} &
\makecell{\textbf{Parameters}\\\textbf{(M)} $\boldsymbol{\downarrow}$} &
\makecell{\textbf{FLOPs}\\\textbf{(G)} $\boldsymbol{\downarrow}$} &
\makecell{\textbf{Model Size}\\\textbf{(MB)} $\boldsymbol{\downarrow}$} &
\makecell{\textbf{Memory}\\\textbf{(MB)} $\boldsymbol{\downarrow}$} &
\textbf{FPS} $\boldsymbol{\uparrow}$ \\

\midrule


\multicolumn{10}{@{}l@{}}{\textbf{Two-Stage Detectors}} \\
\addlinespace[1pt]

\hspace*{0.8em}RoI Transformer (R50)
& 0.9001
& 0.5536
& 0.7710
& 0.5229
& 55.25
& 56.91
& 210.97
& 531.30
& $23.71 \pm 0.44$ \\

\hspace*{0.8em}RoI Transformer (R18)
& 0.8988
& 0.5502
& 0.7687
& 0.5178
& 42.18
& 43.39
& 160.95
& 478.00
& $27.28 \pm 0.55$ \\

\hspace*{0.8em}Oriented R-CNN (R50)
& \underline{0.9003}
& \textbf{0.5672}
& \textbf{0.7870}
& \textbf{0.5361}
& 41.35
& 53.30
& 157.94
& 376.19
& $30.61 \pm 0.53$ \\

\hspace*{0.8em}R-Faster R-CNN (R50)
& 0.8887
& 0.4647
& 0.7297
& 0.4832
& 41.35
& 53.10
& 157.94
& 478.64
& $31.98 \pm 0.58$ \\

\hspace*{0.8em}Oriented R-CNN (R18)
& \textbf{0.9004}
& 0.5609
& 0.7734
& \underline{0.5309}
& 28.28
& 39.78
& 107.92
& 324.56
& $32.84 \pm 0.38$ \\

\hspace*{0.8em}R-Faster R-CNN (R18)
& 0.8884
& 0.4611
& 0.7214
& 0.4809
& 28.28
& 39.58
& 107.92
& 423.96
& $36.88 \pm 0.92$ \\

\midrule


\multicolumn{10}{@{}l@{}}{\textbf{One-Stage Detectors}} \\
\addlinespace[1pt]

\hspace*{0.8em}S$^2$ANet (R50)
& 0.8971
& 0.4682
& 0.7409
& 0.4868
& 38.76
& 49.05
& 148.06
& 317.35
& $22.15 \pm 0.40$ \\

\hspace*{0.8em}R-RetinaNet (R50)
& 0.8728
& 0.4016
& 0.6698
& 0.4379
& 36.35
& 52.39
& 138.87
& 308.15
& $32.91 \pm 0.33$ \\

\hspace*{0.8em}R-FCOS (R50)
& 0.8904
& 0.5322
& 0.7560
& 0.5115
& 32.12
& 51.55
& 122.71
& 292.00
& $33.91 \pm 0.39$ \\

\hspace*{0.8em}RTMDet-m
& 0.8916
& 0.5145
& 0.7422
& 0.5090
& 24.66
& 25.01
& 104.40
& 140.95
& $37.19 \pm 0.63$ \\

\hspace*{0.8em}S$^2$ANet (R18)
& 0.8965
& 0.4635
& 0.7281
& 0.4776
& 22.20
& 36.11
& 84.73
& 229.37
& $27.42 \pm 0.26$ \\

\hspace*{0.8em}R-RetinaNet (R18)
& 0.8707
& 0.4006
& 0.6694
& 0.4363
& 19.79
& 39.46
& 75.54
& 220.43
& $36.35 \pm 0.39$ \\

\hspace*{0.8em}R-FCOS (R18)
& 0.8899
& 0.5260
& 0.7556
& 0.5099
& 19.10
& 38.84
& 72.88
& 217.53
& $38.46 \pm 0.43$ \\

\hspace*{0.8em}RTMDet-s
& 0.8864
& 0.4812
& 0.7357
& 0.5022
& 8.86
& \underline{9.44}
& 38.38
& \underline{65.00}
& $\underline{43.82 \pm 0.56}$ \\

\hspace*{0.8em}RTMDet-tiny
& 0.8861
& 0.4696
& 0.7203
& 0.4919
& \underline{4.87}
& \textbf{5.14}
& \underline{21.19}
& \textbf{42.81}
& $\boldsymbol{45.76 \pm 0.99}$ \\

\midrule


\multicolumn{10}{@{}l@{}}{\textbf{Ours}} \\
\addlinespace[1pt]

\hspace*{0.8em}\textbf{RoI Transformer-slim $^\dagger$}
& 0.8957
& 0.5419
& 0.7629
& 0.5172
& 6.56
& 13.00
& 25.38
& 172.45
& $31.08 \pm 0.32$ \\

\hspace*{0.8em}\textbf{Oriented R-CNN-slim $^\dagger$}
& 0.8988
& \underline{0.5643}
& \underline{0.7736}
& 0.5298
& \textbf{3.40}
& 10.73
& \textbf{12.99}
& 165.38
& $34.82 \pm 0.42$ \\

\bottomrule

\end{tabularx}%
}

\end{table}

For RoI Transformer-slim, the~strongest competing method varies across the
reported metrics. On~SSDD, DTKD improves AP$_{50}$ over MGD by
{0.07\%}. It also improves AP$_{75}$, mAP$_{50:75}$, and~mAP$_{50:95}$ over PKD by
{0.69\%, 0.42\%, and~1.42\%}, respectively.
On RSDD-SAR, DTKD improves AP$_{50}$ over PKD by
{0.02\%}, while exceeding CWD in AP$_{75}$,
mAP$_{50:75}$, and~mAP$_{50:95}$ by
{3.27\%, 0.73\%, and~2.43\%}, respectively.
Here, the~gains are again most apparent in AP$_{75}$ and
mAP$_{50:95}$, further indicating that DTKD is particularly effective under
stricter localization~criteria.

\textls[-15]{Across both student architectures and datasets, DTKD consistently
{achieves the highest mean performance} for every reported metric.
The consistent improvements, particularly under stricter rotated IoU
thresholds, support the effectiveness of proposal-aligned prediction transfer
and complementary teacher guidance for heavily compressed two-stage oriented
detectors.}
\vspace{-3pt}

\begin{table}[H]
\footnotesize
\setlength{\tabcolsep}{14pt}
\caption{Comparison 
with knowledge distillation methods on SSDD and RSDD-SAR. Values are reported as mean $\pm$ standard deviation over three independent~runs. The best results are shown in bold, and the second-best results are
underlined.}
\label{tab:kd_comparison_ssdd_rsdd}

\begin{tabularx}{\textwidth}{rrcccc}
\toprule
\textbf{Data} & \textbf{Method}
& \textbf{AP$\boldsymbol{_{50}}$}
& \textbf{AP$\boldsymbol{_{75}}$}
& \textbf{mAP$\boldsymbol{_{50:75}}$}
& \textbf{mAP$\boldsymbol{_{50:95}}$} \\
\midrule

\multicolumn{6}{@{}l@{}}{{Oriented R-CNN-slim} 
} \\

\multirow{4}{*}{SSDD}
& CWD~\cite{shu2021cwd}
& $\underline{0.9072 \pm 0.0003}$
& $\underline{0.6742 \pm 0.0015}$
& $\underline{0.8445 \pm 0.0009}$
& $\underline{0.5894 \pm 0.0007}$ \\

& PKD~\cite{cao2022pkd}
& $0.9064 \pm 0.0002$
& $0.6679 \pm 0.0013$
& $0.8270 \pm 0.0008$
& $0.5811 \pm 0.0006$ \\

& MGD~\cite{yang2022mgd}
& $0.9059 \pm 0.0002$
& $0.6572 \pm 0.0011$
& $0.8251 \pm 0.0007$
& $0.5660 \pm 0.0005$ \\

& \textbf{DTKD (ours)}
& $\boldsymbol{0.9079 \pm 0.0003}$
& $\boldsymbol{0.6947 \pm 0.0016}$
& $\boldsymbol{0.8489 \pm 0.0010}$
& $\boldsymbol{0.6009 \pm 0.0007}$ \\

\addlinespace[2pt]

\multirow{4}{*}{RSDD}
& CWD~\cite{shu2021cwd}
& $\underline{0.8978 \pm 0.0002}$
& $\underline{0.5588 \pm 0.0011}$
& $\underline{0.7718 \pm 0.0006}$
& $\underline{0.5249 \pm 0.0005}$ \\

& PKD~\cite{cao2022pkd}
& $0.8975 \pm 0.0002$
& $0.5547 \pm 0.0010$
& $0.7692 \pm 0.0005$
& $0.5180 \pm 0.0004$ \\

& MGD~\cite{yang2022mgd}
& $0.8961 \pm 0.0003$
& $0.5532 \pm 0.0016$
& $0.7671 \pm 0.0010$
& $0.5201 \pm 0.0008$ \\

& \textbf{DTKD (ours)}
& $\boldsymbol{0.8986 \pm 0.0003}$
& $\boldsymbol{0.5633 \pm 0.0015}$
& $\boldsymbol{0.7730 \pm 0.0009}$
& $\boldsymbol{0.5293 \pm 0.0007}$ \\


\bottomrule
\end{tabularx} 
\end{table}

\begin{table}[H]\ContinuedFloat
\caption{\textit{Cont.}}
\footnotesize
\setlength{\tabcolsep}{14pt}
\begin{tabularx}{\textwidth}{rrcccc}
\toprule
\textbf{Data} & \textbf{Method}
& \textbf{AP$\boldsymbol{_{50}}$}
& \textbf{AP$\boldsymbol{_{75}}$}
& \textbf{mAP$\boldsymbol{_{50:75}}$}
& \textbf{mAP$\boldsymbol{_{50:95}}$} \\
\midrule

\multicolumn{6}{@{}l@{}}{RoI Transformer-slim} \\

\multirow{4}{*}{SSDD}
& CWD~\cite{shu2021cwd}
& $0.9062 \pm 0.0003$
& $0.6658 \pm 0.0014$
& $0.8407 \pm 0.0009$
& $0.5808 \pm 0.0006$ \\

& PKD~\cite{cao2022pkd}
& $0.9070 \pm 0.0003$
& $\underline{0.6692 \pm 0.0018}$
& $\underline{0.8424 \pm 0.0010}$
& $\underline{0.5829 \pm 0.0007}$ \\

& MGD~\cite{yang2022mgd}
& $\underline{0.9072 \pm 0.0002}$
& $0.6549 \pm 0.0014$
& $0.8268 \pm 0.0007$
& $0.5663 \pm 0.0006$ \\

& \textbf{DTKD (ours)}
& $\boldsymbol{0.9078 \pm 0.0003}$
& $\boldsymbol{0.6738 \pm 0.0014}$
& $\boldsymbol{0.8459 \pm 0.0009}$
& $\boldsymbol{0.5912 \pm 0.0007}$ \\

\addlinespace[2pt]

\multirow{4}{*}{RSDD}
& CWD~\cite{shu2021cwd}
& $0.8953 \pm 0.0002$
& $\underline{0.5256 \pm 0.0011}$
& $\underline{0.7579 \pm 0.0006}$
& $\underline{0.5053 \pm 0.0005}$ \\

& PKD~\cite{cao2022pkd}
& $\underline{0.8956 \pm 0.0003}$
& $0.5213 \pm 0.0015$
& $0.7570 \pm 0.0009$
& $0.5016 \pm 0.0007$ \\

& MGD~\cite{yang2022mgd}
& $0.8951 \pm 0.0002$
& $0.5155 \pm 0.0012$
& $0.7506 \pm 0.0007$
& $0.5044 \pm 0.0005$ \\

& \textbf{DTKD (ours)}
& $\boldsymbol{0.8958 \pm 0.0002}$
& $\boldsymbol{0.5428 \pm 0.0012}$
& $\boldsymbol{0.7634 \pm 0.0007}$
& $\boldsymbol{0.5176 \pm 0.0005}$ \\

\bottomrule
\end{tabularx}

\end{table}
\unskip

\subsection{Distillation Ablation~Study}

We conduct ablation studies on SSDD and RSDD-SAR to examine the effects of
Rotated Proposal Alignment (RPA), the~homogeneous main teacher, and~the
heterogeneous auxiliary teacher. As~shown
in~Table \ref{tab:ablation_ssdd_rsdd}, five configurations are evaluated for
each compressed student. These comprise direct fine-tuning, dual-teacher
distillation without RPA, RPA with only the auxiliary teacher, RPA with only
the main teacher, and~the full DTKD framework. The~student architectures and
training settings are kept~unchanged.

We first assess the role of RPA using direct fine-tuning as the reference.
When both teachers are introduced without RPA, Oriented R-CNN-slim shows an
unchanged AP$_{50}$ and a marginal 0.04\% gain in mAP$_{50:75}$ on SSDD,
while AP$_{75}$ and mAP$_{50:95}$ decrease by 0.01\% and 0.52\%,
respectively. On~RSDD-SAR, its AP$_{50}$, AP$_{75}$, mAP$_{50:75}$, and~mAP$_{50:95}$ decrease by 0.30\%, 0.92\%, 0.44\%, and~0.50\%.
The effect is more pronounced for RoI Transformer-slim, whose four metrics
decrease by 0.11\%, 3.08\%, 0.23\%, and~1.28\% on SSDD and by 1.54\%,
0.76\%, 1.12\%, and~1.35\% on RSDD-SAR. Distillation without RPA therefore provides no consistent benefit, suggesting that unmatched teacher and student proposals introduce noise into prediction-level~distillation.

The auxiliary teacher becomes beneficial when RPA is enabled. Relative to
direct fine-tuning, the~configuration using RPA and the auxiliary teacher
increases AP$_{50}$ by 0.08\% on SSDD but decreases it by 0.02\% on
RSDD-SAR for Oriented R-CNN-slim. Across the two datasets, AP$_{75}$ improves
by 0.70\% and 0.72\%, mAP$_{50:75}$ by 0.12\% and 0.32\%, and~mAP$_{50:95}$ by 0.52\% and 0.17\%, respectively. For~RoI
Transformer-slim, AP$_{50}$ increases by 0.02\% on SSDD and remains
essentially unchanged on RSDD-SAR. The~corresponding gains are 0.05\% and
0.88\% in AP$_{75}$, 1.86\% and 0.28\% in mAP$_{50:75}$, and~1.95\%
and 1.19\% in mAP$_{50:95}$. Thus, although~the AP$_{50}$ changes are
minor, the~auxiliary teacher improves AP$_{75}$ and both mAP metrics in every
student--dataset~setting.

Compared with the configuration using RPA and the auxiliary teacher, the~configuration using RPA and the main teacher improves all reported metrics.
For Oriented R-CNN-slim, the~gains on SSDD and RSDD-SAR are 0.04\% and
0.04\% in AP$_{50}$, 0.44\% and 0.23\% in AP$_{75}$, 1.91\% and
0.04\% in mAP$_{50:75}$, and~1.73\% and 0.74\% in mAP$_{50:95}$,
respectively. For~RoI Transformer-slim, the~corresponding gains are 0.06\%
and 0.09\%, 0.05\% and 1.17\%, 0.21\% and 0.37\%, and~0.90\% and
1.15\%. These results indicate that the architecture-matched main teacher
provides stronger individual supervision than the auxiliary~teacher.

Finally, relative to the RPA-enabled main-teacher configuration, adding the
auxiliary teacher further improves the stricter localization metrics. For~Oriented R-CNN-slim, AP$_{50}$ decreases by 0.01\% on SSDD and increases by
0.06\% on RSDD-SAR, while AP$_{75}$ improves by 1.88\% and 0.57\%,
mAP$_{50:75}$ by 0.22\% and 0.10\%, and~mAP$_{50:95}$ by 1.04\% and
0.40\%, respectively. For~RoI Transformer-slim, AP$_{50}$ decreases by
0.01\% and 0.12\%, whereas AP$_{75}$ increases by 1.65\% and 1.44\%,
mAP$_{50:75}$ by 0.46\% and 0.09\%, and~mAP$_{50:95}$ by 1.01\% and
0.06\%. The~AP$_{50}$ variations remain within 0.12\%, while AP$_{75}$ and
both mAP metrics improve in every setting. Overall, RPA enables reliable
prediction transfer, the~main teacher provides stronger individual
guidance, and~the auxiliary teacher contributes complementary knowledge when
both teachers are~used.

\begin{table}[H]
\footnotesize
\setlength{\tabcolsep}{10pt}
\caption{Ablation study of the proposed distillation design on SSDD and RSDD-SAR.
The checkmark indicates that the corresponding training component or supervision source is used.
Best and second-best results within each student model and dataset are shown in
\textbf{bold} and \underline{underlined}, respectively.}
\label{tab:ablation_ssdd_rsdd}

\begin{tabularx}{\textwidth}{llccccccc}
\toprule
\textbf{Student} &
\textbf{Data} &
\textbf{RPA} &
\makecell{\textbf{Main}\\\textbf{Teacher}} &
\makecell{\textbf{Aux.}\\\textbf{Teacher}} &
\textbf{AP$\boldsymbol{_{50}}$} &
\textbf{AP$\boldsymbol{_{75}}$} &
\textbf{mAP$\boldsymbol{_{50:75}}$} &
\textbf{mAP$\boldsymbol{_{50:95}}$} \\
\midrule

\multirow{10}{*}{\makecell{Oriented\\R-CNN-slim}}
& \multirow{5}{*}{SSDD}
&  &  &  & 0.9069 & 0.6739 & 0.8300 & 0.5815 \\
&
&  & $\checkmark$ & $\checkmark$
& 0.9069 & 0.6738 & 0.8303 & 0.5785 \\
&
& $\checkmark$ &  & $\checkmark$
& 0.9076 & 0.6786 & 0.8310 & 0.5845 \\
&
& $\checkmark$ & $\checkmark$ &
& \textbf{0.9080} & \underline{0.6816} & \underline{0.8469} & \underline{0.5946} \\
&
& $\checkmark$ & $\checkmark$ & $\checkmark$
& \underline{0.9079} & \textbf{0.6944} & \textbf{0.8488} & \textbf{0.6008} \\
\cmidrule{2-9}

& \multirow{5}{*}{RSDD}
&  &  &  & 0.8981 & 0.5558 & 0.7700 & 0.5229 \\
&
&  & $\checkmark$ & $\checkmark$
& 0.8954 & 0.5507 & 0.7666 & 0.5203 \\
&
& $\checkmark$ &  & $\checkmark$
& 0.8979 & 0.5598 & 0.7725 & 0.5238 \\
&
& $\checkmark$ & $\checkmark$ &
& \underline{0.8983} & \underline{0.5611} & \underline{0.7728} & \underline{0.5277} \\
&
& $\checkmark$ & $\checkmark$ & $\checkmark$
& \textbf{0.8988} & \textbf{0.5643} & \textbf{0.7736} & \textbf{0.5298} \\

\midrule

\multirow{10}{*}{\makecell{RoI\\Transformer-slim}}
& \multirow{5}{*}{SSDD}
&  &  &  & 0.9070 & 0.6613 & 0.8243 & 0.5685 \\
&
&  & $\checkmark$ & $\checkmark$
& 0.9060 & 0.6409 & 0.8224 & 0.5612 \\
&
& $\checkmark$ &  & $\checkmark$
& 0.9072 & 0.6616 & 0.8396 & 0.5796 \\
&
& $\checkmark$ & $\checkmark$ &
& \textbf{0.9077} & \underline{0.6619} & \underline{0.8414} & \underline{0.5848} \\
&
& $\checkmark$ & $\checkmark$ & $\checkmark$
& \underline{0.9076} & \textbf{0.6728} & \textbf{0.8453} & \textbf{0.5907} \\
\cmidrule{2-9}

& \multirow{5}{*}{RSDD}
&  &  &  & \underline{0.8960} & 0.5234 & 0.7573 & 0.5050 \\
&
&  & $\checkmark$ & $\checkmark$
& 0.8822 & 0.5194 & 0.7488 & 0.4982 \\
&
& $\checkmark$ &  & $\checkmark$
& \underline{0.8960} & 0.5280 & 0.7594 & 0.5110 \\
&
& $\checkmark$ & $\checkmark$ &
& \textbf{0.8968} & \underline{0.5342} & \underline{0.7622} & \underline{0.5169} \\
&
& $\checkmark$ & $\checkmark$ & $\checkmark$
& 0.8957 & \textbf{0.5419} & \textbf{0.7629} & \textbf{0.5172} \\

\bottomrule
\end{tabularx}

\end{table}

\subsection{Auxiliary Teacher Architecture~Analysis}

\label{sec:aux_teacher_analysis}

{To provide a post-hoc analysis of the effect of auxiliary teacher architecture,
we compare R-Faster R-CNN (R50) and RoI Transformer (R50) as auxiliary
teachers for Oriented R-CNN-slim on the SSDD test set, while the student,
main teacher, and~all other distillation settings are kept unchanged.}

{As shown in~Table \ref{tab:aux_teacher_analysis}, compared with
R-Faster R-CNN, RoI Transformer as the auxiliary teacher improves
AP$_{50}$, AP$_{75}$, mAP$_{50:75}$, and~mAP$_{50:95}$ by
0.08\%, 2.86\%, 1.45\%, and~2.60\%, respectively. The~small gain in
AP$_{50}$ is consistent with the small AP$_{50}$ difference between the
two auxiliary teachers themselves, whereas the larger gains under the
stricter localization metrics correspond to the stronger performance of
RoI Transformer on these metrics. These results show that the choice of
auxiliary teacher affects distillation performance and suggest that
predictive capability is an important consideration when selecting the
heterogeneous auxiliary teacher.}

\begin{table}[H]
\small
\caption{\textls[-20]{Effect of different auxiliary teacher architectures for
Oriented R-CNN-slim on SSDD.}}
\label{tab:aux_teacher_analysis}

\begin{tabularx}{\textwidth}{lCCCC}
\toprule
Auxiliary Teacher &
AP$\boldsymbol{_{50}}$ &
AP$\boldsymbol{_{75}}$ &
mAP$\boldsymbol{_{50:75}}$ &
mAP$\boldsymbol{_{50:95}}$ \\
\midrule

R-Faster R-CNN (R50)
& 0.9072 & 0.6751 & 0.8367 & 0.5856 \\

RoI Transformer (R50)
& 0.9079 
& 0.6944
& 0.8488
& 0.6008 \\

\bottomrule
\end{tabularx}
\end{table}

\subsection{Visualization~Results}

\begin{figure}[H]
\centering

\def\visimgwidth{0.30\textwidth}
\setlength{\tabcolsep}{1.2pt}
\renewcommand{\arraystretch}{0.96}
\small

\makebox[\textwidth][c]{%
\begin{tabular}{@{}ccccc@{}}

& \textbf{GT}
& \textbf{RTMDet-tiny}
& \textbf{RoI Transformer-slim}
& \textbf{Oriented R-CNN-slim} \\[1pt]

\raisebox{8mm}{\rotatebox{90}{\makecell{SSDD\\inshore}}}
&
\includegraphics[width=\visimgwidth]{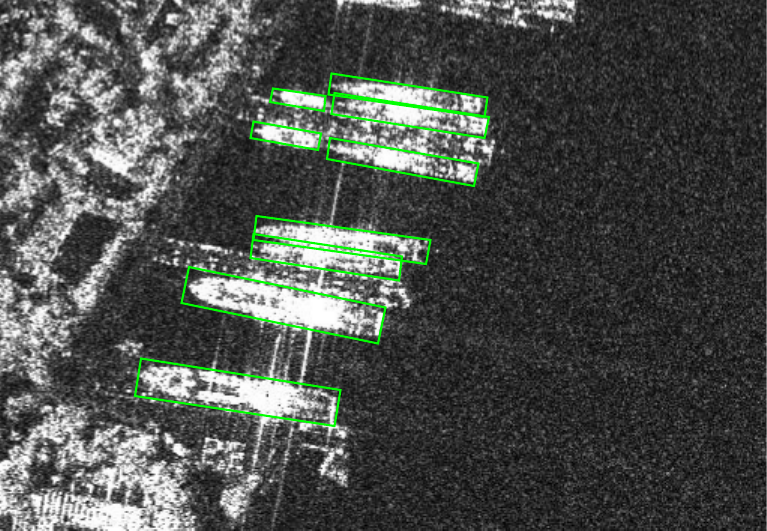}
&
\includegraphics[width=\visimgwidth]{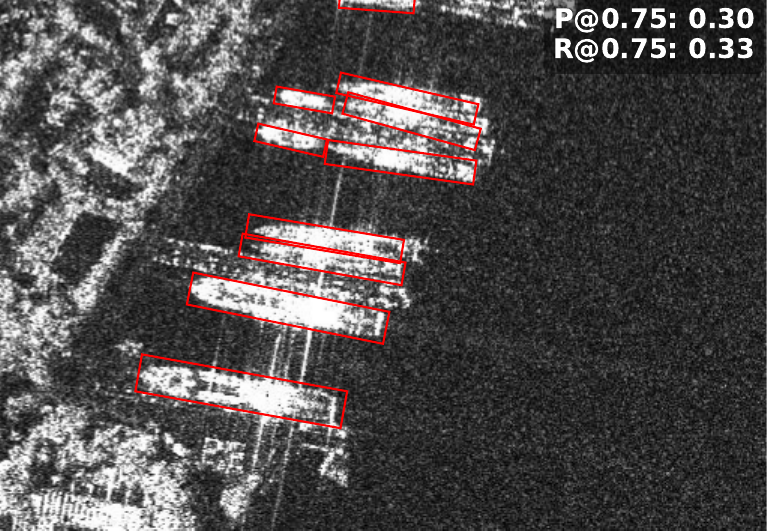}
&
\includegraphics[width=\visimgwidth]{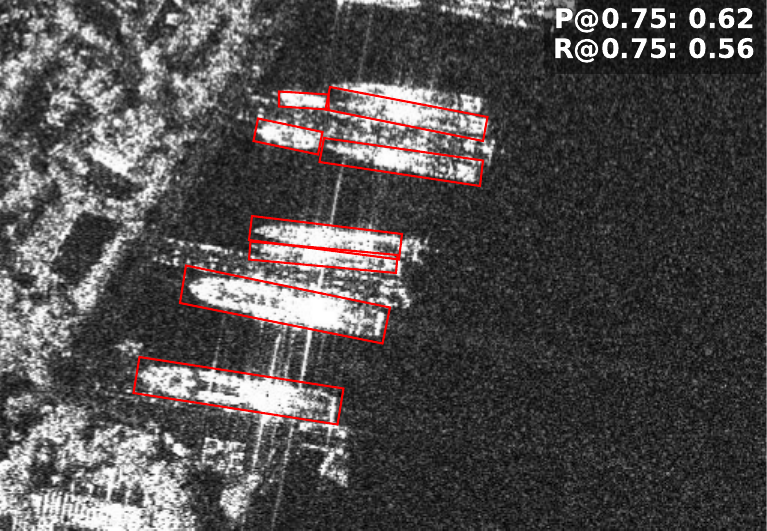}
&
\includegraphics[width=\visimgwidth]{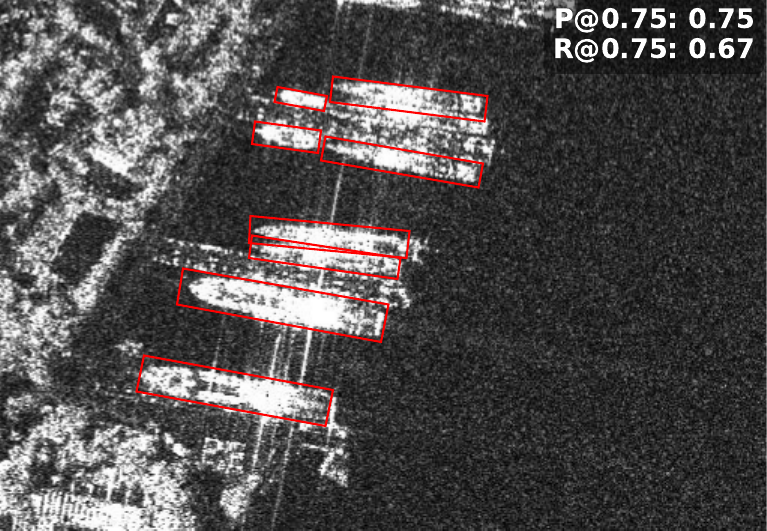}
\\[2pt]

\raisebox{8mm}{\rotatebox{90}{\makecell{SSDD\\offshore}}}
&
\includegraphics[width=\visimgwidth]{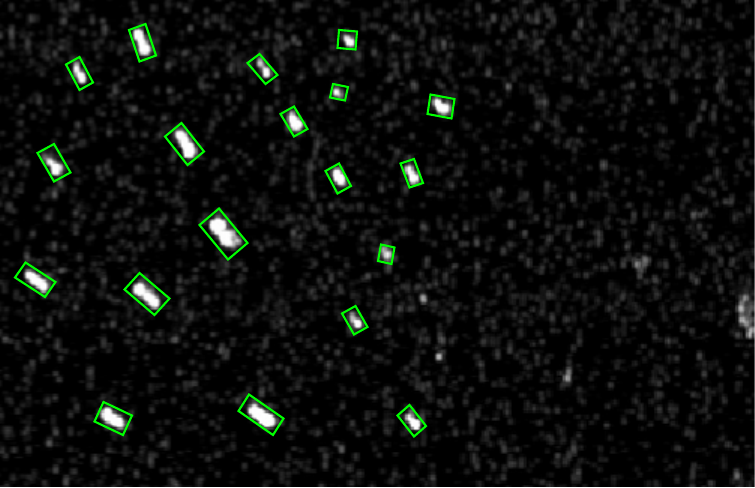}
&
\includegraphics[width=\visimgwidth]{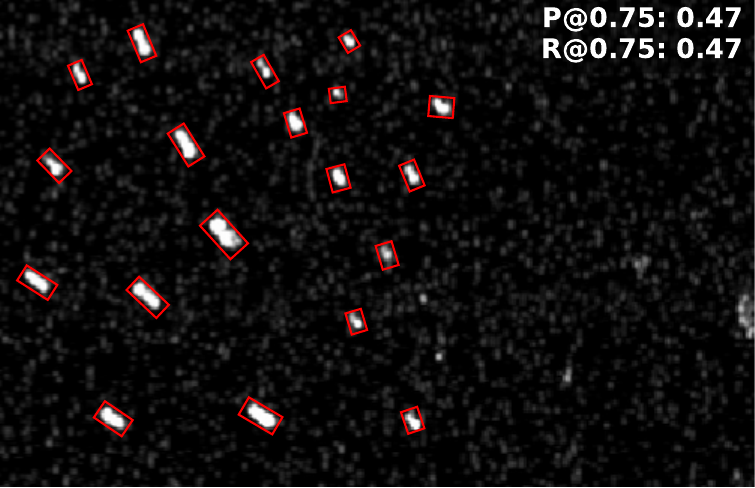}
&
\includegraphics[width=\visimgwidth]{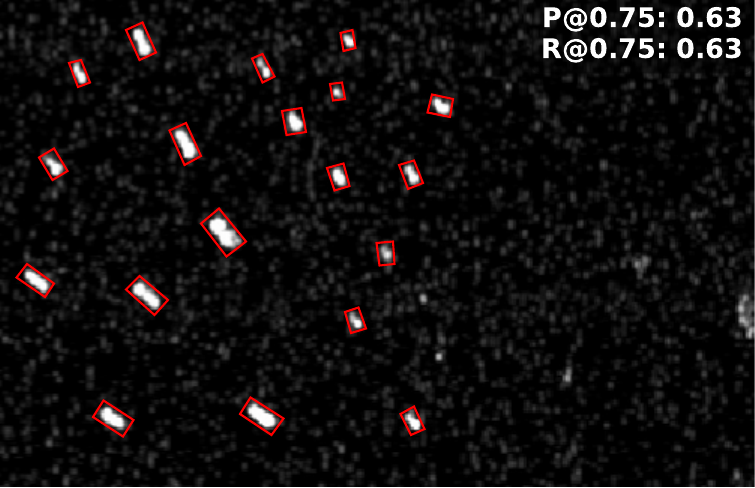}
&
\includegraphics[width=\visimgwidth]{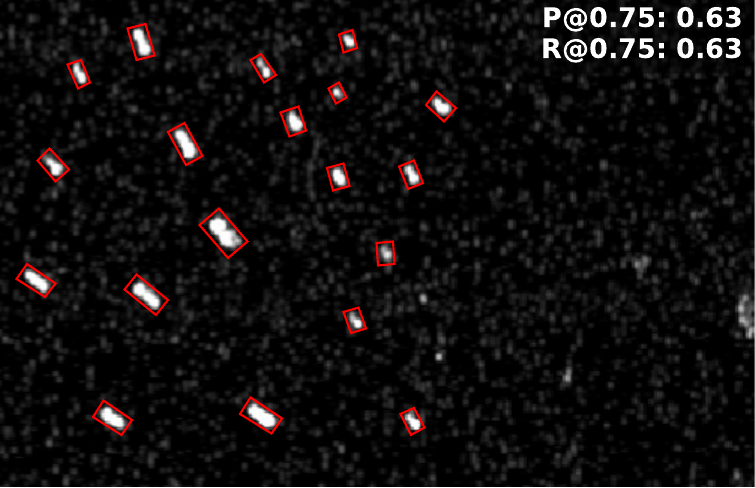}
\\[2pt]

\raisebox{14mm}{\rotatebox{90}{\makecell{RSDD\\offshore}}}
&
\includegraphics[width=\visimgwidth]{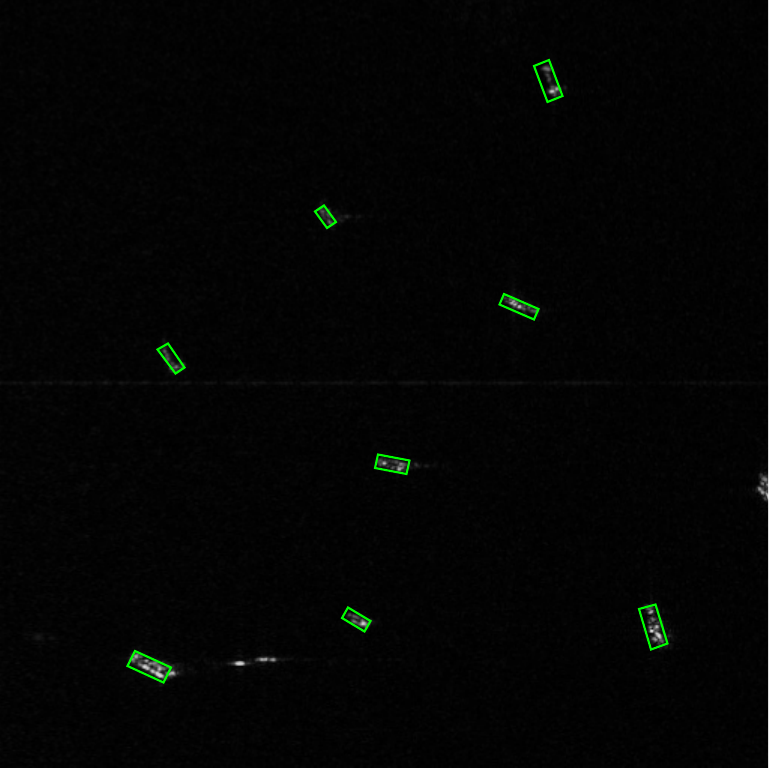}
&
\includegraphics[width=\visimgwidth]{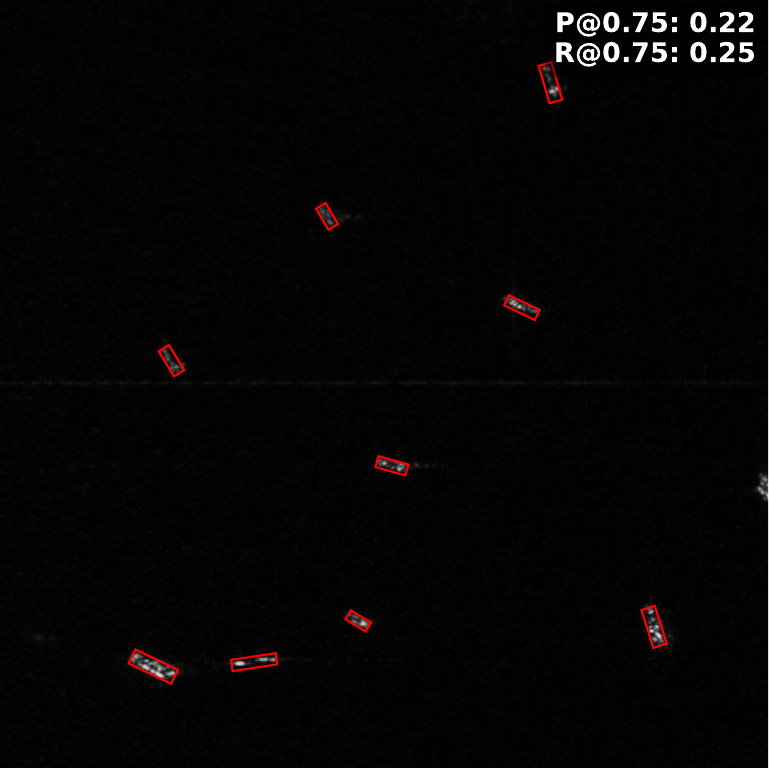}
&
\includegraphics[width=\visimgwidth]{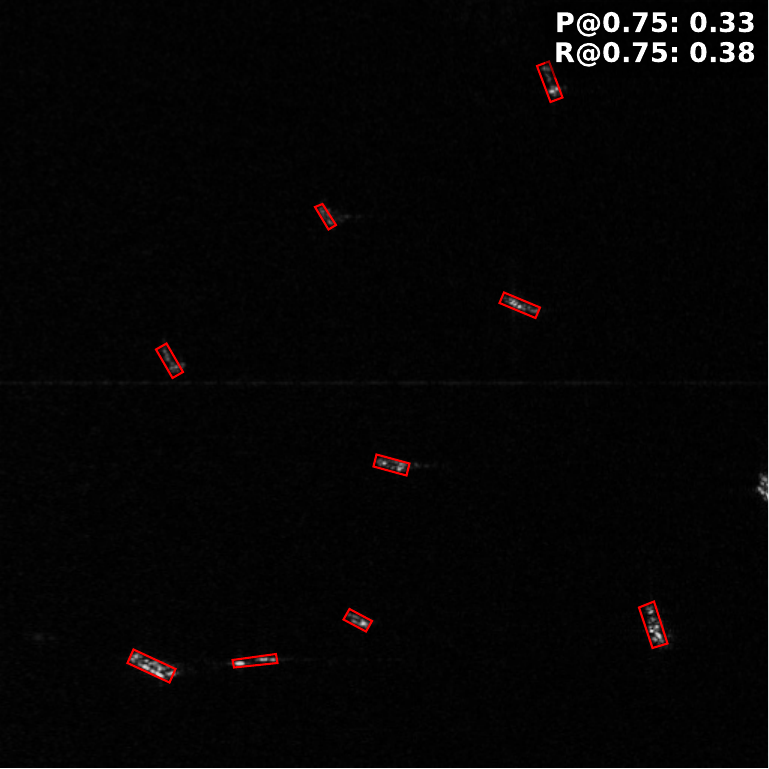}
&
\includegraphics[width=\visimgwidth]{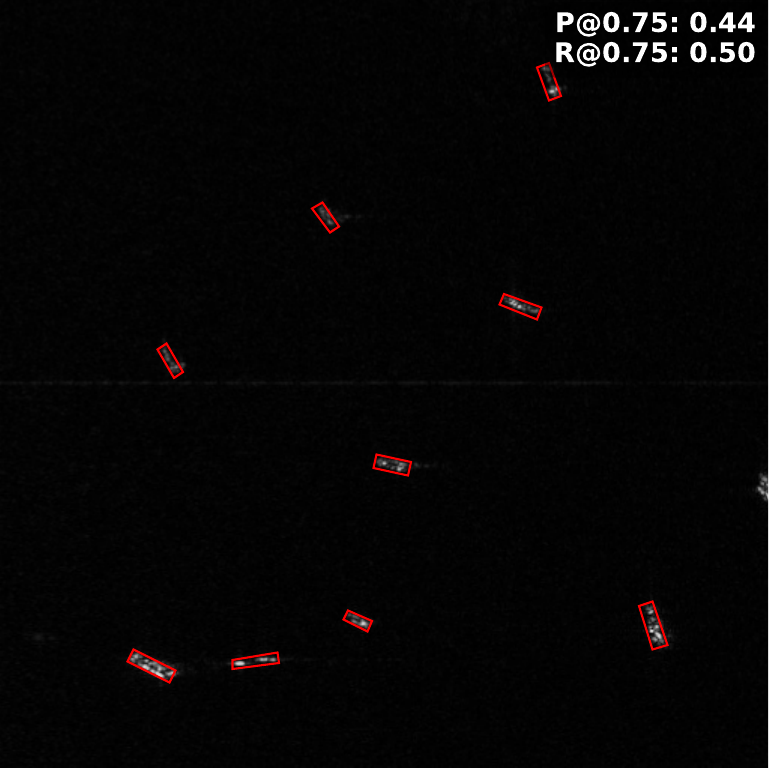}
\\[2pt]

\raisebox{14mm}{\rotatebox{90}{\makecell{RSDD\\inshore}}}
&
\includegraphics[width=\visimgwidth]{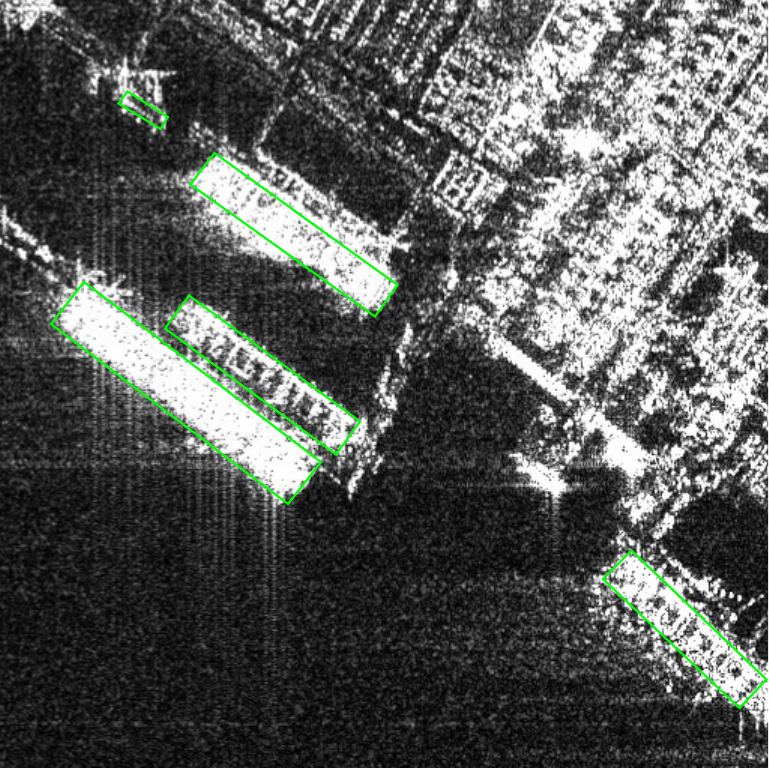}
&
\includegraphics[width=\visimgwidth]{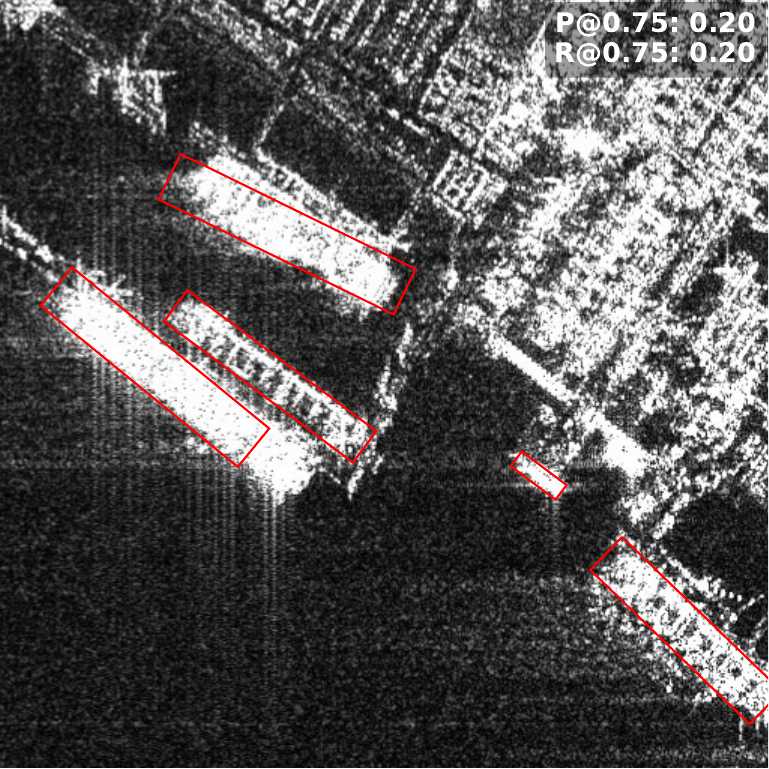}
&
\includegraphics[width=\visimgwidth]{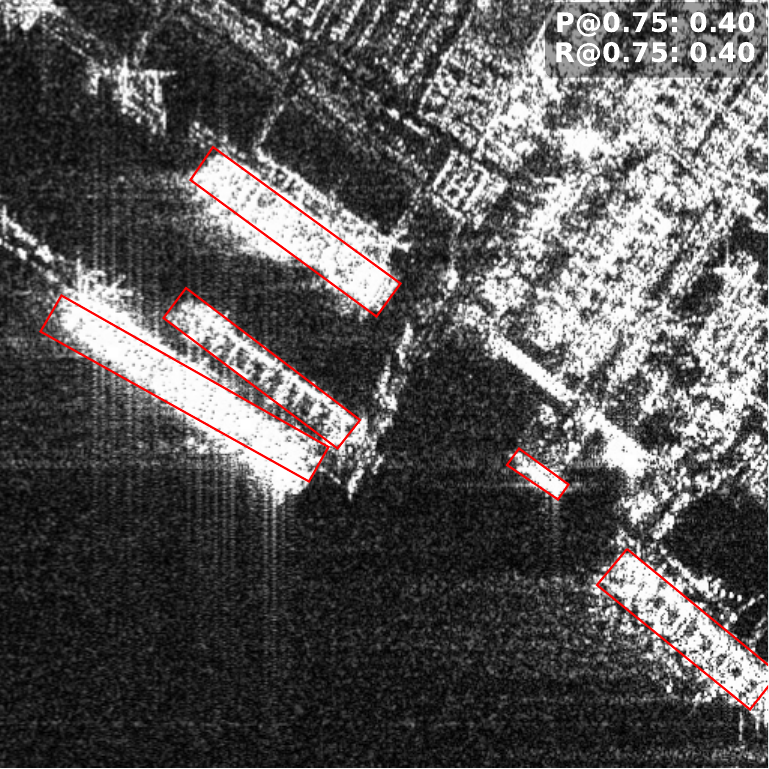}
&
\includegraphics[width=\visimgwidth]{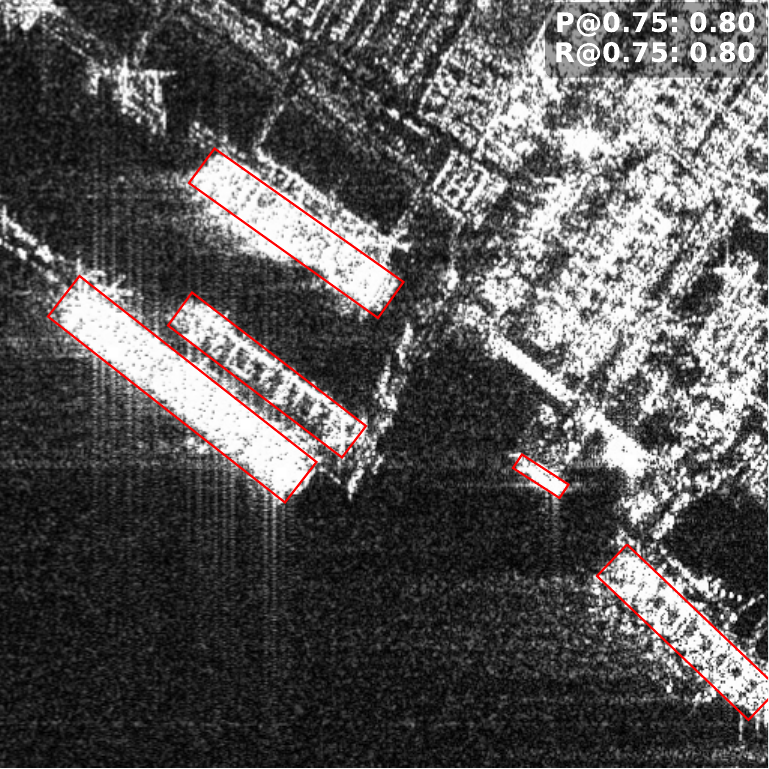}

\end{tabular}%
}

\vspace{1mm}

\caption{
Qualitative detection results on SSDD and RSDD-SAR under inshore and offshore
SAR ship scenarios. The compressed two-stage detectors produce tighter oriented
boxes and fewer localization errors than the lightweight one-stage baseline.
The top-right annotations report precision and recall under the 0.75 IoU
threshold for each prediction image. Green boxes denote ground-truth
annotations, while red boxes denote detection results.
}

\label{fig:visualization_results}

\end{figure}

Figure \ref{fig:visualization_results} presents qualitative detection results on
SSDD and RSDD-SAR under both inshore and offshore SAR ship scenarios. The~examples cover cluttered near-shore backgrounds, sparse offshore scenes, and~densely distributed small ships. Compared with RTMDet-tiny, the~proposed compressed two-stage detectors produce tighter and more stable oriented bounding boxes, with~fewer missed detections and reduced localization deviations. This is particularly evident in near-shore scenes, where coastal structures and strong background scattering can easily interfere with dense one-stage predictions. The~visual results are consistent with the quantitative comparison, confirming that the compressed two-stage detectors better preserve precise rotated localization capability under challenging SAR imaging~conditions.

\section{Discussion}
\label{sec:discussion}
The experimental results demonstrate that DTKDP can compress accurate two-stage
oriented detectors into lightweight models while preserving their localization
advantage for SAR ship detection. Unlike many existing lightweight SAR ship
detectors that are built on compact one-stage dense predictors, DTKDP compresses
the entire two-stage detection pipeline, including the backbone, FPN, RPN,
and~RoI head. The~pruning analyses reveal architecture-dependent sensitivity to
sparsity regularization and distinct gate distributions across detector
components, supporting architecture-specific sparsity settings and
component-specific pruning thresholds. Under comparable parameter budgets,
component-wise pruning consistently outperforms both backbone-only and global
pruning across the two detector architectures and datasets. Together with the
substantial parameter reductions achieved in the FPN, RPN, and~RoI head, these
results support extending structured pruning to the full two-stage detection
pipeline rather than restricting compression to the backbone or using a single
global pruning threshold.

Relative to their full-scale R50 counterparts, the~proposed compressed
detectors retain comparable detection accuracy while substantially reducing
model complexity. {They also generally achieve stronger overall accuracy than
the evaluated one-stage baselines while maintaining competitive inference
efficiency under the tested RTX 3090 configuration.} In particular, both
compressed detectors outperform RTMDet-tiny across all reported accuracy
metrics on SSDD and RSDD-SAR. Although~RTMDet-tiny requires fewer FLOPs and
less peak GPU memory and achieves higher FPS, the~proposed detectors remain
compact while delivering consistently higher accuracy. Overall, these results
demonstrate a favorable accuracy--efficiency~trade-off.

The distillation analyses demonstrate the effectiveness of the dual-teacher
design and the importance of proposal alignment and teacher complementarity.
Compared with representative distillation methods, DTKD consistently achieves
superior detection performance across both student architectures and datasets,
particularly under stricter rotated IoU criteria. The~ablation results further
show that direct knowledge transfer after component-wise pruning is challenging
because changes in the feature representations used by the RPN may lead to
mismatched proposal distributions between the teacher and student. RPA
alleviates this mismatch by evaluating teacher and student predictions in a
shared rotated proposal space. The~dual-teacher design uses the homogeneous
teacher for stable classification and regression guidance and the heterogeneous
teacher for complementary classification cues. The~ablation results also show
that the dual-teacher configuration provides additional improvements over the
corresponding single-teacher configurations. The~teacher complementarity
analysis further shows that the two teacher architectures exhibit different
strengths across ship scales and aspect ratios, while the auxiliary-teacher
analysis indicates that the choice of heterogeneous teacher can affect
distillation performance.

Despite these advantages, several limitations remain. First, {the compressed
two-stage detectors still exhibit lower inference speed and higher peak GPU
memory consumption than the most efficient one-stage baselines.
Although pruning substantially reduces the parameter count and FLOPs,
these metrics do not fully capture the runtime and memory costs of a two-stage
detection pipeline. Candidate proposals must first be generated and filtered
before RoI-level prediction, after~which the selected proposals undergo RoI
feature extraction and prediction. This stage-wise dependency limits end-to-end
parallelism, while proposal filtering, RoI feature extraction, and~per-proposal
prediction incur additional memory access and execution overheads that are not
fully captured by FLOPs. As~a result, reductions in FLOPs do not necessarily
produce proportional gains in FPS. Similarly, parameter pruning primarily
reduces the memory required for model weights, whereas peak inference memory
also depends on intermediate feature maps, proposal tensors, RoI features,
and~temporary buffers. Because~these intermediate representations are not
reduced in direct proportion to the parameter count, the~reduction in peak
memory consumption is therefore more limited. These overheads may become more
pronounced on resource-constrained embedded platforms with limited memory
bandwidth and parallel processing capability.}
Potential deployment-oriented improvements include hardware-aware optimization,
sparse proposal selection, and~operator fusion to reduce inference latency,
together with memory-efficient RoI processing to lower peak GPU
memory~consumption.

{Second, the~current experiments are conducted on SSDD and RSDD-SAR.
Although these datasets cover multiple sensors, spatial resolutions,
and~maritime scenes, broader validation under lower-resolution imagery,
different radar frequencies, polarizations, and~sea-clutter conditions is still
needed. In particular, our group previously developed the NovaSAR Automated
Ship Target Recognition (NASTaR) dataset~\cite{hosseiny2026nastar} from
relatively lower-resolution NovaSAR S-band imagery with AIS-associated ship
labels. NASTaR was originally designed for ship-type classification, and~we are
currently extending it with ship-level bounding-box annotations to support
object detection. Once completed, the~extended dataset will provide an
additional testbed for evaluating DTKDP under lower-resolution S-band imagery
and different maritime conditions. The~NASA--ISRO Synthetic Aperture Radar
(NISAR) mission, which provides L- and S-band SAR observations, also offers a
potential source for future cross-frequency evaluation~\cite{rosen2017nisar}.
Such evaluations will help assess the generalizability of DTKDP beyond the
imaging conditions represented by the current benchmarks.}

{Finally, two limitations remain in the current distillation framework.
The dual-teacher formulation adopts fixed distillation weights across
proposals. Although~the sensitivity analysis in
Section~\ref{sec:distill_weight_analysis} shows that the distillation benefits
are maintained across a range of fixed weight settings, the teacher
complementarity analysis indicates that the relative strengths of the two
teachers vary with ship scale and aspect ratio. Adaptive teacher weighting has
been explored in multi-teacher knowledge distillation, where teacher
contributions are dynamically adjusted according to input-dependent confidence
information~\cite{yu2023adaptive}. Motivated by this direction, adaptive
teacher weighting based on teacher confidence, target scale, or~aspect ratio
will be investigated to provide more targeted knowledge transfer. In~addition,
the~auxiliary teacher is currently used only for classification distillation.
Geometry-normalized regression targets will be explored to extend
heterogeneous teacher distillation from classification to localization.}

\section{Conclusions}
\label{sec:conclusions}

This paper presents DTKDP, a~dual-teacher knowledge distillation and pruning
framework for lightweight two-stage oriented SAR ship detection. Learnable
gates are introduced into convolutional, normalization, and~linear layers to
jointly prune convolutional channels and RoI-head neurons, enabling
component-wise compression of the backbone, feature pyramid, proposal module,
and detection head. Rotated Proposal Alignment performs prediction-level
distillation on shared teacher-generated rotated proposals, while the
dual-teacher strategy combines classification and regression guidance from a
homogeneous main teacher with complementary classification knowledge from a
heterogeneous \mbox{auxiliary~teacher.}

{The pruning analyses show that component-wise pruning provides a better
accuracy--complexity trade-off than backbone-only and global pruning under
comparable parameter budgets.}
Experiments on SSDD and RSDD-SAR show that DTKDP reduces the parameter counts
of Oriented R-CNN (R50) and RoI Transformer (R50) by 87.5--91.8\% and their
FLOPs by 75.6--79.9\%. Despite these substantial complexity reductions,
the~resulting Oriented R-CNN-slim and RoI Transformer-slim models retain
accuracy close to that of their full-scale counterparts, with~relative changes
across AP$_{50}$, AP$_{75}$, mAP$_{50:75}$, and~mAP$_{50:95}$ ranging from
a 2.38\% decrease to a 0.65\% improvement.
{Compared with RTMDet-tiny, the~compressed detectors improve all four accuracy
metrics on both datasets, with~relative gains ranging from 0.52\% to 27.55\%.
RTMDet-tiny requires fewer FLOPs and less peak GPU memory and achieves higher
FPS, whereas the proposed models provide stronger rotated localization
performance under the tested RTX 3090 configuration.}
DTKD also consistently outperforms representative distillation methods,
including CWD, PKD, and~MGD, across both student architectures and datasets.
{Overall, these results demonstrate that full-pipeline compression of
two-stage oriented detectors can provide a favorable accuracy--efficiency
trade-off, particularly when strict rotated localization is prioritized.}

{Future work will focus on hardware-level validation of the final slim models
on representative resource-constrained platforms, including batch-1 latency,
throughput, peak memory usage, and~energy or power consumption. Potential
deployment improvements include hardware-aware optimization, sparse proposal
selection, operator fusion, and~memory-efficient RoI processing. Broader
cross-sensor and cross-frequency validation under diverse SAR imaging
conditions will also be pursued. Adaptive teacher weighting and
geometry-normalized regression targets will be explored as extensions to the
current distillation framework.}

\vspace{+30pt}

\reftitle{References}

\end{document}